\documentclass[11pt,oneside,fleqn,reqno,titlepage]{article}
\usepackage{amsmath}
\usepackage{amssymb}
\usepackage{amsthm}
\usepackage{natbib} \citeindextrue
\usepackage{comment}
\usepackage{url}
\usepackage{bbm}
\usepackage{bm}
\usepackage{multirow}
\usepackage{relsize}
\usepackage[pdftex]{graphicx}
\usepackage{color}

\newcommand{\doublespace}{\addtolength{\baselineskip}{.25\baselineskip}}
\newcommand{\doublespacetwo}{\addtolength{\baselineskip}{1.0\baselineskip}}
\newcommand{\singlespace}{\addtolength{\baselineskip}{-.5\baselineskip}}

\newcounter{stepno}

\newcommand{\E}{\mathbb{E}}
\newcommand{\bn}{\begin{eqnarray}}
\newcommand{\en}{\end{eqnarray}}
\newcommand{\bns}{\begin{eqnarray*}}
\newcommand{\ens}{\end{eqnarray*}}

\newcommand{\defvarbegin}{\begin{quotation}\vspace{-15pt}\begin{tabbing}}
\newcommand{\defvarend}  {\end{tabbing}\vspace{-10pt}\end{quotation}}

\newcommand{\bnarray}{\begin{equation}\begin{array}{rcll}}
\newcommand{\enarray}{\end{array}\end{equation}}

\newcommand{\textwrap}{\parbox[t]{5.5in}}

\newcommand{\barr}{\begin{array}}
\newcommand{\earr}{\end{array}}

\newcounter{cnum}

\newcommand{\beginalg}{\setcounter{stepno}{1}
                \begin{list}{\bf Step~\arabic{stepno}}
                         {\usecounter{stepno}\settowidth{\labelwidth}{\bf Step~9m}
                \addtolength{\leftmargin}{2\parindent}}
                }
\newcommand{\eg}{\end{list}}

\def \define{\begin{quote}\begin{itemize}}
\def \enddefine{\end{itemize}\end{quote}}

\newlength{\boxedparwidth} 
  {\begin{center} \begin{tabular}{|@{\hspace{.15in}}c@{\hspace{.15in}}|}
                 \hline \\ \begin{minipage}[t]{\boxedparwidth}
                 \setlength{\parindent}{0.0in}}%
   {\end{minipage} \\ \\ \hline \end{tabular} \end{center}}
\newcounter{example}
\def \rhooperator{\mathlarger{\mathlarger{\rho}}}

\newcommand{\argmax}{{\rm arg}\max}

\def \ptilde{{\tilde p}}

\def \xtilde{{\tilde x}}

\def \Ctilde{{\tilde C}}

\def \Etilde{{\tilde E}}

\def \Rtilde{{\tilde R}}
\def \Stilde{{\tilde S}}

\def \Wtilde{{\tilde W}}
\def \Xtilde{{\tilde X}}

\def \phat{\hat p}
\def \qhat{\hat q}

\def \Dhat{\hat D}
\def \Ehat{\hat E}

\def \What{\widehat W}

\def \thetabar{\bar \theta}

\def \hbar{\bar h}

\def \pbar{\bar p}

\def \Cbar{\bar C}

\def \Fbar{\bar F}
\def \Qbar{\bar Q}

\def \Vbar{\bar V}

\def \mubar{\bar \mu}
\def \sigmabar{\bar \sigma}

\def \Acal{{\cal A}}

\def \Fcal{{\cal F}}

\def \Qcal{{\cal Q}}

\def \Scal{{\cal S}}

\def \Xcal{{\cal X}}

\begin{document}

\title{From Reinforcement Learning to Optimal Control: A unified framework for sequential decisions}
\author{Warren B. Powell \\ Department of Operations Research and Financial Engineering \\ Princeton University \\}
\date{\today}

\maketitle

\clearpage

\begin{abstract}
There are over 15 distinct communities that work in the general area of sequential decisions and information, often referred to as decisions under uncertainty or stochastic optimization.  We focus on two of the most important fields: stochastic optimal control, with its roots in deterministic optimal control, and reinforcement learning, with its roots in Markov decision processes.  Building on prior work, we describe a unified framework that covers all 15 different communities, and note the strong parallels with the modeling framework of stochastic optimal control.  By contrast, we make the case that the modeling framework of reinforcement learning, inherited from discrete Markov decision processes, is quite limited.  Our framework (and that of stochastic control) is based on the core problem of optimizing over policies.  We describe four classes of policies that we claim are universal, and show that each of these two fields have, in their own way, evolved to include examples of each of these four classes.
\end{abstract}

\clearpage
\pagestyle{empty}
{\small
{\singlespace
\contentsline {section}{\numberline {1}Introduction}{1}%
\contentsline {section}{\numberline {2}The communities of sequential decisions}{3}%
\contentsline {section}{\numberline {3}Stochastic optimal control vs. reinforcement learning}{5}%
\contentsline {subsection}{\numberline {3.1}Stochastic control}{5}%
\contentsline {subsection}{\numberline {3.2}Reinforcement learning}{8}%
\contentsline {subsection}{\numberline {3.3}A critique of the MDP modeling framework}{13}%
\contentsline {subsection}{\numberline {3.4}Bridging optimal control and reinforcement learning}{14}%
\contentsline {section}{\numberline {4}The universal modeling framework}{16}%
\contentsline {subsection}{\numberline {4.1}Dimensions of a sequential decision model}{17}%
\contentsline {subsection}{\numberline {4.2}State variables}{20}%
\contentsline {subsection}{\numberline {4.3}Objective functions}{21}%
\contentsline {subsection}{\numberline {4.4}Notes}{23}%
\contentsline {section}{\numberline {5}Energy storage illustration}{24}%
\contentsline {subsection}{\numberline {5.1}A basic energy storage problem}{25}%
\contentsline {subsection}{\numberline {5.2}With a time-series price model}{27}%
\contentsline {subsection}{\numberline {5.3}With passive learning}{27}%
\contentsline {subsection}{\numberline {5.4}With active learning}{28}%
\contentsline {subsection}{\numberline {5.5}With rolling forecasts}{28}%
\contentsline {subsection}{\numberline {5.6}Remarks}{30}%
\contentsline {section}{\numberline {6}Designing policies}{30}%
\contentsline {subsection}{\numberline {6.1}Policy search}{31}%
\contentsline {subsection}{\numberline {6.2}Lookahead approximations}{32}%
\contentsline {subsection}{\numberline {6.3}Hybrid policies}{35}%
\contentsline {subsection}{\numberline {6.4}Remarks}{36}%
\contentsline {subsection}{\numberline {6.5}Stochastic control, reinforcement learning, and the four classes of policies}{37}%
\contentsline {section}{\numberline {7}Policies for energy storage}{40}%
\contentsline {section}{\numberline {8}Extension to multiagent systems}{42}%
\contentsline {section}{\numberline {9}Observations}{43}%
\contentsline {section}{References}{45}%
}
}
\clearpage

\setcounter{page}{1}
\pagestyle{plain}

\doublespace

\section{Introduction}

There is a vast range of problems that consist of the sequence: decisions, information, decisions, information, $\ldots$. Application areas span engineering, business, economics, finance, health, transportation, and energy.  It encompasses active learning problems that arise in the experimental sciences, medical decision making, e-commerce, and sports.  It also includes iterative algorithms for stochastic search, as well as two-agent games and multiagent systems. In fact, we might claim that virtually any human enterprise will include instances of sequential decision problems.

Given the diversity of problem domains, it should not be a surprise that a number of communities have emerged to address the problem of making decisions over time to optimize some metric.  The reason that so many communities exist is a testament to the variety of problems, but it also hints at the many methods that are needed to solve these problems.  As of this writing, there is not a single method that has emerged to solve all problems.  In fact, it is fair to say that all the methods that have been proposed are {\it fragile}: relatively modest changes can invalidate a theoretical result, or increase run times by orders of magnitude.

In \cite{PowellEJOR2019}, we present a unified framework for all sequential decision problems.  This framework consists of a mathematical model (that draws heavily from the framework used widely in stochastic control), which requires optimizing over {\it policies} which are functions for making decisions given what we know at a point in time (captured by the state variable).

The significant advance of the unified framework is the identification of four (meta)classes of policies that encompass all the communities.  In fact, whereas the solution approach offered by each community is fragile, we claim that the four classes are universal: any policy proposed for any sequential decision problem will consist of one of these four classes, and possibly a hybrid.

The contribution of the framework is to raise the visibility of all of the communities.  Instead of focusing on a specific solution approach (for example, the use of Hamilton-Jacobi-Bellman (HJB) equations, which is one of the four classes), the framework encourages people to consider all four classes, and then to design policies that are best suited to the characteristics of a problem.

This chapter is going to focus attention on two specific communities: stochastic optimal control, and reinforcement learning.  Stochastic optimal control emerged in the 1950's, building on what was already a mature community for deterministic optimal control that emerged in the early 1900's and has been adopted around the world.  Reinforcement learning, on the other hand, emerged in the 1990's building on the foundation of Markov decision processes which was introduced in the 1950's (in fact, the first use of the term ``stochastic optimal control'' is attributed to Bellman, who invented Markov decision processes).  Reinforcement learning emerged from computer science in the 1980's, and grew to prominence in 2016 when it was credited with solving the Chinese game of Go using AlphaGo.

We are going to make the following points:
\begin{itemize}
\item Both communities have evolved from a core theoretical/algorithm result based on Hamilton-Jacobi-Bellman equations, transitioning from exact results (that were quite limited), to the use of algorithms based on approximating value functions/cost-to-go functions/Q-factors, to other strategies that do not depend on HJB equations.  We will argue that each of the fields is in the process of recognizing all four classes of policies.
\item We will present and contrast the canonical modeling frameworks for stochastic control and reinforcement learning (adopted from Markov decision processes).  We will show that the framework for stochastic control is very flexible and scalable to real applications, while that used by reinforcement learning is limited to a small problem class.
\item We will present a universal modeling framework for sequential decision analytics (given in \cite{PowellEJOR2019}) that covers any sequential decision problem.  The framework draws heavily from that used by stochastic control, with some minor adjustments.  While not used by the reinforcement learning community, we will argue that it is used implicitly.  In the process, we will dramatically expand the range of problems that can be viewed as either stochastic control problems, or reinforcement learning problems.
\end{itemize}

We begin our presentation in section \ref{sec:communities} with an overview of the different communities that work on sequential decisions under uncertainty, along with a list of major problem classes.  Section \ref{sec:optimalcontrolvsreinforcementlearning} presents a side-by-side comparison of the modeling frameworks of stochastic optimal control and reinforcement learning.

Section \ref{sec:framework} next presents our universal framework (taken from \cite{PowellEJOR2019}), and argues that a) it covers all 15+ fields (presented in section \ref{sec:communities}) dealing with sequential decisions and uncertainty, b) it draws heavily from the standard model of stochastic optimal control, and c) the framework of reinforcement learning, inherited from discrete Markov decision processes, has fundamental weaknesses that limit its applicability to a very narrow classes of problems.  We then illustrate the framework using an energy storage problem in section \ref{sec:energystorage}; this application offers tremendous richness, and allows us to illustrate the flexibility of the framework.

The central challenge of our modeling framework involves optimizing over policies, which represents our point of departure with the rest of the literature, since it is standard to pick a class of policy in advance.  However, this leaves open the problem of how to search over policies.  In section \ref{sec:policies} we present four (meta)classes of policies which, we claim, are universal, in that {\it any} approach suggested in the literature (or in practice) is drawn from one of these four classes, or a hybrid of two or more.  Section \ref{sec:policiesforenergystorage} illustrates all four classes, along with a hybrid, using the context of our energy storage application.  These examples will include hybrid resource allocation/active learning problems, along with the overlooked challenge of dealing with rolling forecasts.

Section \ref{sec:multiagent} briefly discusses how to use the framework to model multiagent systems, and notes that this vocabulary provides a fresh perspective on partially observable Markov decision processes.  Then, section \ref{sec:observations} concludes the chapter with a series of observations about reinforcement learning, stochastic optimal control, and our universal framework.

\section{The communities of sequential decisions}
\label{sec:communities}

The list of potential applications of sequential decision problems is virtually limitless.  Below we list a number of major application domains.  Ultimately we are going to model all of these using the same framework.
\begin{description}
  \item[Discrete problems] These are problems with discrete states and discrete decisions (actions), such as stochastic shortest path problems.
  \item[Control problems] These span controlling robots, drones, rockets and submersibles, where states are continuous (location and velocity) as are controls (forces).  Other examples include determining optimal dosages of medications, or continuous inventory (or storage) problems that arise in finance and energy.
  \item[Dynamic resource allocation problems] Here we are typically managing inventories (retail products, food, blood, energy, money, drugs), typically over space and time.  It also covers discrete problems such as dynamically routing vehicles, or managing people or machines.  It would also cover planning the movements of robots and drones (but not how to do it). The scope of ``dynamic resource allocation problems'' is almost limitless.
  \item[Active learning problems] This includes any problem that involves learning, and where decisions affect what information is collected (laboratory experiments, field experiments, test marketing, computer simulations, medical testing). It spans multiarmed bandit problems, e-commerce (bidding, recommender systems), black-box simulations, and simulation-optimization.
  \item[Hybrid learning/resource allocation problems] This would arise if we are managing a drone that is collecting information, which means we have to manage a physical resource while running experiments which are then used to update beliefs.  Other problems are laboratory science experiments with setups (this is the physical resource), collecting public health information from field technicians, and any experimental learning setting with a budget constraint.
  \item[Stochastic search] This includes both derivative-based and derivative-free stochastic optimization.
  \item[Adversarial games] This includes any two-player (or multiplayer) adversarial games.  It includes pricing in markets where price affects market behavior, and military applications.
  \item[Multiagent problems] This covers problems with multiple decision-makers who might be competing or cooperating.  They might be making the same decisions (but spatially distributed), or making different decisions that interact (as arises in supply chains, or where different agents play different roles but have to work together).
\end{description}

Given the diversity of problems, it should not be surprising that a number of different research communities have evolved to address them, each with their own vocabulary and solution methods, creating what we have called the ``jungle of stochastic optimization'' (\cite{Powell2014}, see also \url{jungle.princeton.edu}).  A list of the different communities that address the problem of solving sequential decision-information problems might be:

\singlespace
\begin{itemize}
  \item Stochastic search (derivative-based)
  \item Ranking and selection (derivative-free)
  \item (Stochastic) optimal control
  \item Markov decision processes/dynamic programming
  \item Simulation-optimization
  \item Optimal stopping
  \item Model predictive control
  \item Stochastic programming
  \item Chance-constrained programming
  \item Approximate/adaptive/neuro-dynamic programming
  \item Reinforcement learning
  \item Robust optimization
  \item Online computation
  \item Multiarmed bandits
  \item Active learning
  \item Partially observable Markov decision processes
\end{itemize}
\doublespacetwo
{\it Each} of these communities is supported by at least one book and over a thousand papers.

Some of these fields include problem classes that can be described as static: make decision, see information (possibly make one more decision), and then the problem stops (stochastic programming and robust optimization are obvious examples).  However, all of them include problems that are fully sequential, consisting of sequences of decision, information, decision, information, $\ldots$, over a finite or infinite horizon.  The focus of this chapter is on fully sequential problems.

Several of the communities offer elegant theoretical frameworks that lead to optimal solutions for specialized problems (Markov decision processes and optimal control are two prominent examples).  Others offer asymptotically optimal algorithms: derivative-based and certain derivative-free stochastic optimization problems, simulation-optimization, and certain instances of approximate dynamic programming and reinforcement learning.  Still others offer theoretical guarantees, often in the form of regret bounds (that is, bounds on how far the solution is from optimal).


We now turn our attention to focus on the fields of stochastic optimal control and reinforcement learning.




\section{Stochastic optimal control vs. reinforcement learning}
\label{sec:optimalcontrolvsreinforcementlearning}
There are numerous communities that have contributed to the broad area of modeling and solving sequential decision problems, but there are two that stand out: optimal control (which laid the foundation for stochastic optimal control), and Markov decision processes, which provided the analytical foundation for reinforcement learning.  Although these fields have intersected at different times in their history, today they offer contrasting frameworks which, nonetheless, are steadily converging to common solution strategies.

We present the modeling frameworks of (stochastic) optimal control and reinforcement learning (drawn from Markov decision processes), which are polar opposites.  Given the growing popularity of reinforcement learning, we think it is worthwhile to compare and contrast these frameworks.  We then present our own universal framework which spans all the fields that deal with any form of sequential decision problems.  The reader will quickly see that our framework is quite close to that used by the (stochastic) optimal control community, with a few adjustments.

\subsection{Stochastic control}
The field of optimal control enjoys a long and rich history, as evidenced by the number of popular books that have been written focusing on deterministic control, including \cite{lewis2012}, \cite{Ki98},  and \cite{Stengel94}.  There are also a number of books on stochastic control (see \cite{sethi2019}, \cite{Nisio2014}, \cite{Sontag1998}, \cite{stengel1986}, \cite{Bertsekas1978}, \cite{Kushner1971}) but these tend to be mathematically more advanced.

Deterministic optimal control problems are typically written
\bn
\min_{u_0, \ldots, u_T} \sum_{t=0}^{T-1} L_t(x_t,u_t) + L_T(x_T), \label{eq:canonicaldeterministiccontrol}
\en
where $x_t$ is the state at time $t$, $u_t$ is the control (that is, the decision) and $L_t(x_t,u_t)$ is a loss function with terminal loss $L_T(x_T)$. The state $x_t$ evolves according to
\bn
x_{t+1} = f_t(x_t,u_t),  \label{eq:deterministiccontroltransition}
\en
where $f_t(x_t,u_t)$ is variously known as the transition function, system model, plant model (as in chemical or power plant), plant equation, and transition law.  We write the control problem in discrete time, but there is an extensive literature where this is written in continuous time, and the transition function is written
\bns
\dot{x}_t = f_t(x_t,u_t).
\ens

The most common form of a stochastic control problem simply introduces additive noise to the transition function given by
\bn
x_{t+1} = f_t(x_t,u_t) + w_t,  \label{eq:stochasticcontroltransition}
\en
where $w_t$ is random at time $t$.  This odd notation arose because of the continuous time formulation, where $w_t$ would be disturbances (such as wind pushing against an aircraft) between $t$ and $t+dt$.  The introduction of the uncertainty means that the state variable $x_t$ is a random variable when we are sitting at time 0.  Since the control $u_t$ is also a function of the state, this means that $u_t$ is also a random variable.  Common practice is to then take an expectation of the objective function in equation \eqref{eq:canonicaldeterministiccontrol}, which produces
\bn
\min_{u_0, \ldots, u_T} \E \left\{\sum_{t=0}^{T-1} L_t(x_t,u_t) + L_T(x_T)\right\}, \label{eq:canonicalstochasticcontrol}
\en
which has to be solved subject to the constraint in equation \eqref{eq:stochasticcontroltransition}. This is problematic, because we have to interpret \eqref{eq:stochasticcontroltransition} recognizing that $x_t$ and $u_t$ are random since they depend on the sequence $w_0, \ldots, w_{t-1}$.

In the deterministic formulation of the problem, we are looking for an optimal control vector $u^*_0, \ldots, u^*_T$.  When we introduce the random variable $w_t$, then the controls need to be interpreted as functions that depend on the information available at time $t$. Mathematicians handle this by saying that ``$u_t$ must be $\Fcal_t$-measurable'' which means, in plain English, that the control $u_t$ is a function (not a variable) which can only depend on information up through time $t$. This leaves us the challenge of finding this function.


We start by relaxing the constraint in \eqref{eq:stochasticcontroltransition} and add it to the objective function, giving us
\bn
\min_{u_0, \ldots, u_T} \E \left\{\sum_{t=0}^{T-1} L_t(x_t,u_t) + L_T(x_T) + \lambda_t(f(x_t,u_t) + w_t - x_{t+1})\right\}.  \label{eq:canonicalstochasticcontrolrelaxed}
\en
where $(\lambda_t)_{t=0}^T$ is a vector of dual variables (known as costate variables in the controls community).  Assuming that $\E w_t = 0$, then $w_t$ drops out of the objective function.

The next step is that we restrict our attention to quadratic loss functions given by
\bn
L_t(x_t,u_t) = (x_t)^T Q_t x_t + (u_t)^T R_t u_t, \label{eq:quadraticloss}
\en
where $Q_t$ and $R_t$ are a set of known matrices.  This special case is known as linear-quadratic regulation (or LQR).

With this special structure, we turn to the Hamilton-Jacobi equations (often called the Hamilton-Jacobi-Bellman equations) where we solve for the ``cost-to-go'' function $J_t(x_t)$ using
\bn
J_t(x_t) = \min_{u} \big(L_t(x_t,u) + \E_{w} J_{t+1}(f(x_t,u,w))\big). \label{eq:HJBequation}
\en
$J_t(x_t)$ is the value of being in state $x_t$ at time $t$ and following an optimal policy from time $t$ onward.  In the language of reinforcement learning, $J_t(x_t)$ is known as the value function, and is written $V_t(S_t)$.

For the special case of the quadratic objective function in \eqref{eq:quadraticloss}, it is possible to solve the Hamilton-Jacobi-Bellman equations analytically and show that the optimal control as a function of the state is given by
\bn
u_t = K_t x_t, \label{eq:linearcontrol}
\en
where $K_t$ is a matrix that depends on $Q_t$ and $R_t$.

Here, $u_t$ is a {\it function} that we refer to as a {\it policy} $\pi$, but is known as a {\it control law} in the controls community.  Some would write \eqref{eq:linearcontrol} as $\pi_t(x_t) = K_t x_t$.  Later we are going to adopt the notation $U^\pi(x_t)$ for writing a policy, where $\pi$ carries information about the structure of the policy.  We note that when the policy depends on the state $x_t$, then the function is, by construction, ``$\Fcal_t$-measurable,'' so we can avoid this terminology entirely.

The linear control law (policy) in equation \eqref{eq:linearcontrol} is very elegant, but it is a byproduct of the special structure, which includes the quadratic form of the objective function (equation \eqref{eq:quadraticloss}), the additive noise (equation \eqref{eq:stochasticcontroltransition}), and the fact that there are no constraints on the controls.  For example, a much more general way of writing the transition function is
\bns
x_{t+1} = f(x_t,u_t,w_t),
\ens
which allows the noise to enter the dynamics in any form.  For example, consider an inventory problem where the state (the inventory level) is governed by
\bns
x_{t+1} = \max\{0, x_t + u_t - w_{t+1}\},
\ens
where $w_{t+1}$ is the random demand for our product.

We are also interested in general state-dependent reward functions which are often written as $g(x_t,u_t)$ (where $g(\cdot)$ stands for gain), as well as the constraints, where we might write
\bns
A_t u_t & = & b_t, \\
    u_t & \geq & 0,
\ens
where $b_t$ (and $A_t$) may contain information from the state variable.

For these more general problems, we cannot compute \eqref{eq:HJBequation} exactly, so the research community has developed a variety of methods for approximating $J_t(x_t)$.  Methods for solving \eqref{eq:HJBequation} approximately have been widely studied in the controls community under names such as heuristic dynamic programming, approximate dynamic programming, neuro-dynamic programming, and adaptive dynamic programming.  However, even this approach is limited to special classes of problems within our universe of sequential decision problems.

Optimal control enjoys a rich history.  Deterministic control dates to the early 1900's, while stochastic control appears to have been first introduced by Bellman in the 1950's (known as the father of dynamic programming).  Some of the more recent books in optimal control are \cite{Ki98}, \cite{stengel1986}, \cite{Sontag1998}, \cite{sethi2019}, and \cite{lewis2012}.  The most common optimal control problems are continuous, low-dimensional and unconstrained.  Stochastic problems are most typically formulated with additive noise.

The field of stochastic control has tended to evolve using the more sophisticated mathematics that has characterized the field.  Some of the most prominent books include \cite{Astrom1970}, \cite{Kushner1971}, \cite{Bertsekas1978}, \cite{YongZhou1999}, \cite{Nisio2014} (note that some of the books on deterministic controls touch on the stochastic case).

We are going to see below that this framework for writing sequential decision problems is quite powerful, even if the classical results (such as the linear control policy) are very limited.  It will form the foundation for our unified framework, with some slight adjustments.


\subsection{Reinforcement learning}
\label{sec:reinforcementlearning}
\begin{figure}[h!]
\center{\includegraphics[width=4.50in]{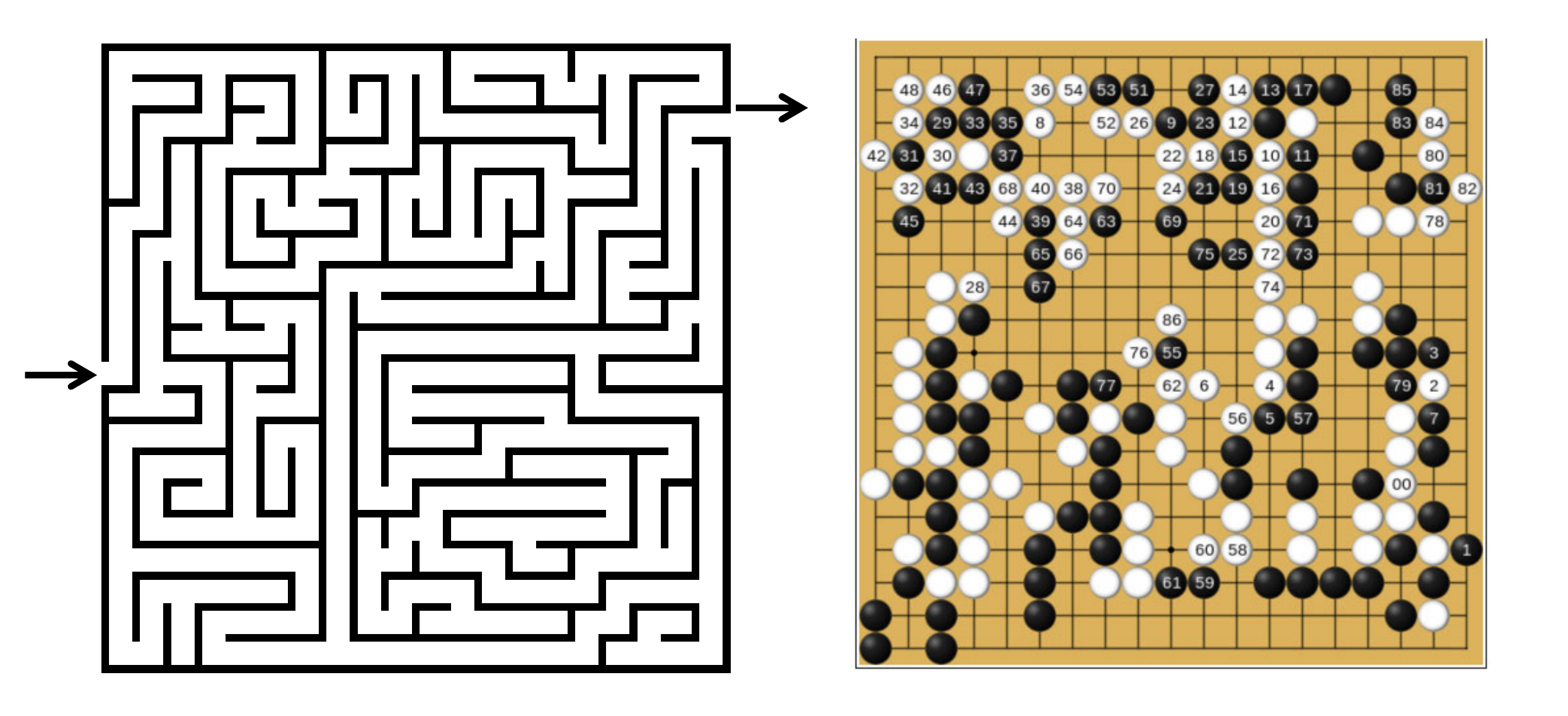}}
\caption{From the mouse-in-the-maze problem, to Chinese Go - a trajectory of reinforcement learning.}
\label{fig:mazeGo}
\end{figure}

The field known as reinforcement learning evolved from early work done by Rich Sutton and his adviser Andy Barto in the early 1980's.  They addressed the problem of modeling the search process of a mouse exploring a maze, developing methods that would eventually help solve the Chinese game of Go, outperforming world masters (figure \ref{fig:mazeGo}).

Sutton and Barto eventually made the link to the field of Markov decision processes and adopted the vocabulary and notation of this field.  The field is nicely summarized in \cite{Puterman05} which can be viewed as the capstone volume on 50 years of research into Markov decision processes, starting with the seminal work of Bellman \citep{Be57}.  \cite{Puterman05}[Chapter 3] summarizes the modeling framework as consisting of the following elements:
\begin{description}
  \item[Decision epochs] $T=1,2, \ldots, N$.
  \item[State space] $\Scal =$ set of (discrete) states.
  \item[Action space] $\Acal =$ action space (set of actions when we are in state $s$).
  \item[Transition matrix] $p(s'|s,a) = $ probability of transitioning to state $s'$ given that we are in state $s$ and take action $a$.
  \item[Reward] $r(s,a) = $ the reward received when we are in state $s$ and take action $a$.
\end{description}
This notation (which we refer to below as the ``MDP formal model'') became widely adopted in the computer science community where reinforcement learning evolved.  It became standard for authors to define a reinforcement learning problem as consisting of the tuple $(\Scal, \Acal, P, r)$ where $P$ is the transition matrix, and $r$ is the reward function.

Using this notation, Sutton and Barto (this work is best summarized in their original volume \cite{Sutton1998}) proposed estimating the value of being in a state $s^n$ and taking an action $a^n$ (at the $nth$ iteration of the algorithm) using
\bn
\qhat^n(s^n,a^n) &=& r(s^n,a^n) + \gamma \max_{a'\in\Acal_{s'}} \Qbar^{n-1}(s',a'), \label{eq:qlearning1} \\
\Qbar^n(s^n,a^n) &=& (1-\alpha_n) \Qbar^{n-1}(s^n,a^n) + \alpha_n \qhat^n(s^n,a^n), \label{eq:qlearning2}
\en
where $\gamma$ is a discount factor and $\alpha_n$ is a smoothing factor that might be called a stepsize (the equation has roots in stochastic optimization) or learning rate.  Equation \eqref{eq:qlearning2} is the core of ``reinforcement learning.''

We assume that when we are in state $s^n$ and take action $a^n$ that we have some way of simulating the transition to a state $s'$.  There are two ways of doing this:
\begin{itemize}
  \item Model-based - We assume we have the transition matrix $P$ and then sample $s'$ from the probability distribution $p(s'|s,a)$.
  \item Model-free - We assume we are observing a physical setting where we can simply observe the transition to state $s'$ without a transition matrix.
\end{itemize}
Sutton and Barto named this algorithmic strategy ``$Q$-learning'' (after the notation).  The appeal of the method is its sheer simplicity.  In fact, they retained this style in their wildly popular book \citep{Sutton1998} which can be read by a high-school student.

Just as appealing is the wide applicability of both the model and algorithmic strategy. Contrast the core algorithmic step described by equations \eqref{eq:qlearning1} - \eqref{eq:qlearning2} to Bellman's equation which is the foundation of Markov decision processes, which requires solving
\bn
V_t(s) = \max_a \big(r(s,a) + \gamma \sum_{s'\in\Scal} p(s'|s,a) V_{t+1}(s')\big),  \label{eq:bellman}
\en
for all states $s\in\Scal$. Equation \eqref{eq:bellman} is executed by setting $V_{T+1}(s) = 0$ for all $s\in\Scal$, and then stepping backward $t=T, T-1, \ldots, 1$ (hence the reason that this is often called ``backward dynamic programming'').  In fact, this version was so trivial that the field focused on the stationary version which is written
\bn
V(s) = \max_a \big(r(s,a) + \gamma \sum_{s'\in\Scal} p(s'|s,a) V(s')\big). \label{eq:bellmansteadystate}
\en
[Side note: The steady state version of Bellman's equation in \eqref{eq:bellmansteadystate} became the default version of Bellman's equation, which explains why the default notation for reinforcement learning does not index variables by time.  By contrast, the default formulation for optimal control is finite time, and variables are indexed by time in the canonical model.]

Equation \eqref{eq:bellmansteadystate} requires solving a system of nonlinear equations to find $V(s)$, which proved to be the foundation for an array of papers with elegant algorithms, where one of the most important is
\bn
V^{n+1}(s) = \max_a \big(r(s,a) + \gamma \sum_{s'\in\Scal} p(s'|s,a) V^n(s')\big).  \label{eq:bellman2}
\en
Equation \eqref{eq:bellman2} is known as value iteration (note the similarity with equation \eqref{eq:bellman}) and is the basis of $Q$-learning (compare to equations \eqref{eq:qlearning1}-\eqref{eq:qlearning2}).

The problem with \eqref{eq:bellman} is that it is far from trivial.  In fact, it is quite rare that it can be computed due to the widely cited ``curse of dimensionality.''  This typically refers to the fact that for most problems, the state $s$ is a vector $s=(s_1, s_2, \ldots, s_K)$.  Assuming that all the states are discrete, the number of states grows exponentially in $K$.  It is for this reason that dynamic programming is widely criticized for suffering from the ``curse of dimensionality.''  In fact, the curse of dimensionality is due purely to the use of lookup table representations of the value function (note that the canonical optimal control model does not do this).

In practice, this typically means that one-dimensional problems can be solved in under a minute; two-dimensional problems might take several minutes (but possibly up to an hour, depending on the dimensionality and the planning horizon); three dimensional problems easily take a week or a month; and four dimensional problems can take up to a year (or more).

In fact, there are actually three curses of dimensionality:  the state space, the action space, and the outcome space.  It is typically assumed that there is a discrete set of actions (think of the roads emanating from an intersection), but there are many problems where decisions are vectors (think of all the ways of assigning different taxis to passengers).  Finally, there are random variables (call them $W$ for now) which might also be a vector.  For example, $W$ might be the set of riders calling our taxi company for rides in a 15 minute period.  Or, it might represent all the attributes of a customer clicking on ads (age, gender, location).

The most difficult computational challenge in equation \eqref{eq:bellman} is finding the one-step transition matrix $P$ with element $p(s'|s,a)$.  This matrix measures $|\Scal| \times |\Scal| \times |\Acal|$ which may already be quite large.  However, consider what it takes to compute just one element.  To show this, we need to steal a bit of notation from the optimal control model, which is the {\it transition function} $f(x,u,w)$.  Using the notation of dynamic programming, we would write this as $f(s,a,w)$.  Assuming the transition function is known, the one-step transition matrix is computed using
\bn
p(s'|s,a) = \E_w\{\mathbbm{1}_{\{s'=f(s,a,w)\}}\}. \label{eq:transitionexpectation}
\en
This is not too hard if the random variable $w$ is scalar, but there are many problems where $w$ is a vector, in which case we encounter the third curse of dimensionality.  There are many other problems where we do not even know the distribution of $w$ (but have a way of observing outcomes).

Now return to the $Q$-learning equations \eqref{eq:qlearning1} - \eqref{eq:qlearning2}.  At no time are we enumerating all the states, although we do have to enumerate all the actions (and the states these actions lead to),  which is perhaps a reason why reinforcement learning is always illustrated in the context of relatively small, discrete action spaces (think of the Chinese game of Go).  Finally, we do not need to take an expectation over the random variable $w$; rather, we just simulate our way from state $s$ to state $s'$ using the transition function $f(s,a,w)$.

We are not out of the woods.  We still have to estimate the value of being in state $s$ and taking action $a$, captured by our $Q$-factors $\Qbar(s,a)$.  If we use lookup tables, this means we need to estimate $\Qbar(s,a)$ for each state that we {\it might} visit, and each action that we {\it might} take, which means we are back to the curse of dimensionality.  However, we can use other approximation strategies:
\begin{itemize}
  \item Lookup tables with hierarchical beliefs - Here we use a family of lookup table models at different levels of aggregation.
  \item Parametric models, which might be linear (in the parameters) or nonlinear. We include shallow neural networks here.  Parametric models transform the dimensionality of problems down to the dimensionality of the parameter vector, but we have to know the parametric form.
  \item Nonparametric models.  Here we include kernel regression, locally parametric, and flexible architectures such as support vector machines and deep neural networks.
\end{itemize}
Not surprisingly, considerable attention has been devoted to different methods for approximating $\Qbar(s,a)$, with recent attention focusing on using deep neural networks.  We will just note that the price of higher-dimensional architectures is that they come with the price of increased training (in fact, dramatically increased training).  A deep neural network might easily require tens of millions of iterations, and yet still may not guarantee high quality solutions.
\begin{figure}[tb]
\begin{center}
    \includegraphics[width=3.5in]{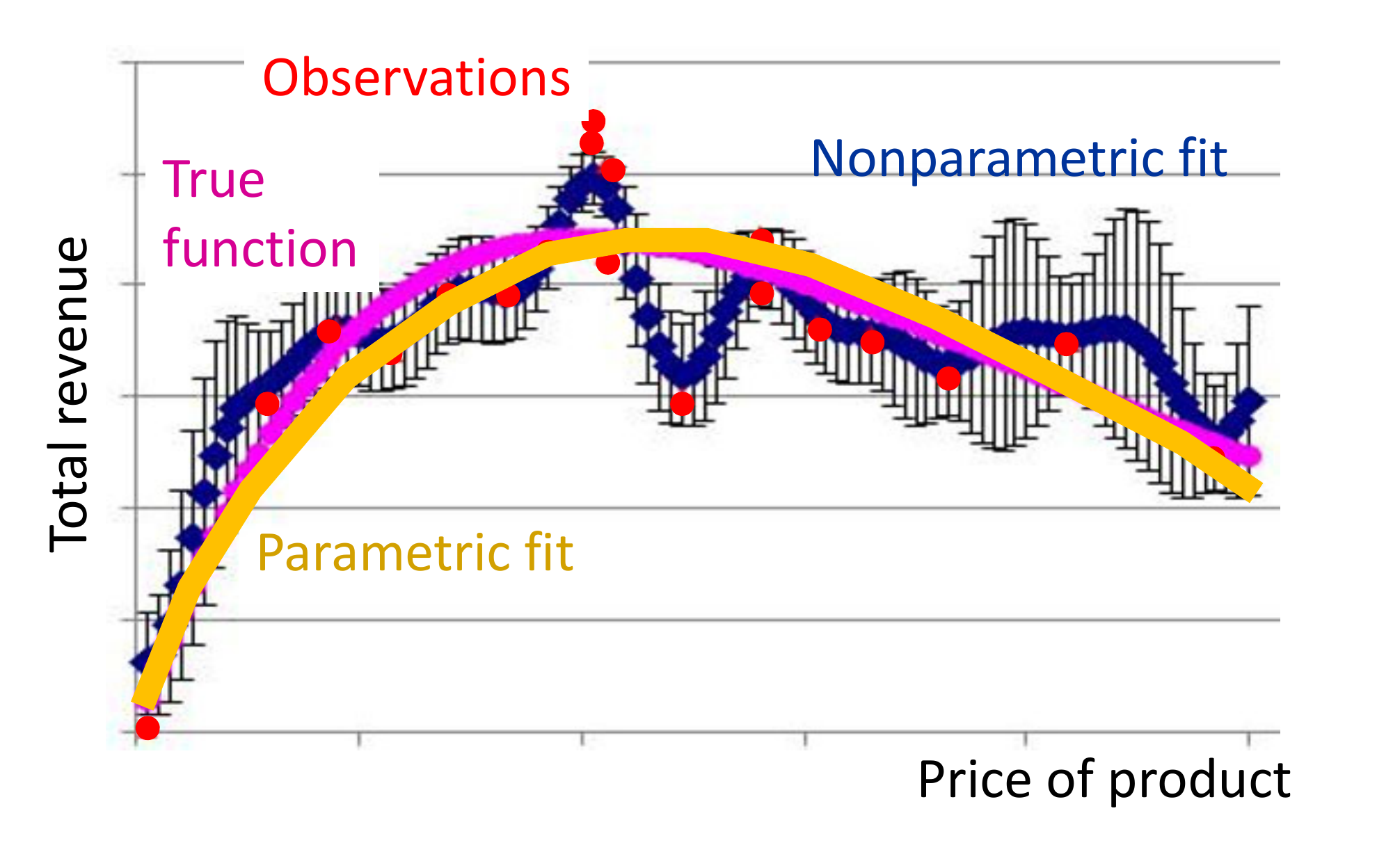}
    \caption{Nonparametric fit (blue line) vs. parametric fit (yellow line), compared to actual (purple line).}
    \label{fig:parametricvsnonparametric}
\end{center}
\end{figure}

The challenge is illustrated in figure \ref{fig:parametricvsnonparametric}, where we have a set of observations (red dots). We try to fit the observations using a nonparametric model (the blue line) which overfits the data, which we believe is a smooth, concave (approximately quadratic) surface.  We would get a reasonable fit of a quadratic function with no more than 10 data points, but any nonparametric model (such as a deep neural network) might require hundreds to thousands of data points (depending on the noise) to get a good fit.

\subsection{A critique of the MDP modeling framework}
For many years, the modeling framework of Markov decision processes lived within the MDP community which consisted primarily of applied probabilists, reflecting the limited applicability of the solution methods.  Reinforcement learning, however, is a field that is exploding in popularity, while still clinging to the classical MDP modeling framework (see \cite{lazaric2019} for a typical example of this).  What is happening, however, is that people doing computational work are adopting styles that overcome the limitations of the discrete MDP framework.  For example, researchers will overcome the problem of computing the one-step transition matrix $p(s'|s,a)$ by saying that they will ``simulate'' the process.  In practice, this means that they are using the transition function $f(s,a,w)$, which means that they have to simulate the random information $w$, without explicitly writing out $f(s,a,w)$ or the model of $w$.  This introduces a confusing gap between the statement of the model and the software that captures and solves the model.

We offer the following criticisms of the classical MDP framework that the reinforcement learning community has adopted:
\begin{itemize}
  \item The MDP/RL modeling framework models state {\it spaces}.  The optimal control framework models state {\it variables}.  We argue that the latter is much more useful, since it more clearly describes the actual variables of the problem.  Consider a problem with discrete states (perhaps with $K$ dimensions). The state space could then be written $\Scal = \Scal_1 \times \cdots \times \Scal_K$ which produces a set of discrete states that we can write $\{1, 2, \ldots, |\Scal|\}$.  If we had a magical tool that could solve discrete Markov decision problems (remember we need to compute the one-step transition matrix), then we do not need to know anything about the state space $\Scal$, but this is rarely the case.  Further, we make the case that just knowing that we have $|\Scal|$ states provides no information about the problem itself, while a list of the variables that make up the state variable (as is done in optimal control) will map directly to the software implementing the model.
  \item Similarly, the MDP/RL community talks about action {\it spaces}, while the controls community uses control {\it variables}.  There is a wide range of problems that are described by discrete actions, where the action space is not too large.  However, there are also many problems where actions are continuous, and often are vector valued.  The notation of an ``action space'' $\Acal$ is simply not useful for vector-valued decisions/controls (the issue is the same as with state spaces).
  \item The MDP modeling framework does not explicitly model the exogenous information process $w_t$.  Rather, it is buried in the one-step transition function $p(s'|s,a)$, as is seen in equation \eqref{eq:transitionexpectation}.  In practical algorithms, we need to simulate the $w_t$ process, so it helps to model the process explicitly.  We would also argue that the model of $w_t$ is a critical and challenging dimension of any sequential decision problem which is overlooked in the canonical MDP modeling framework.
  \item Transition functions, if known, are always computable, since we just have to compute them for a single state, a single action, and a single observation of any exogenous information.  We suspect that this is why the optimal control community adopted this notation.   One-step transition matrices (or one-step transition kernels if the state variable is continuous) are almost never computable.
  \item There is no proper statement of the objective function, beyond the specification of the reward function $r(s,a)$.  There is an implied objective similar to equation \eqref{eq:canonicalstochasticcontrol}, but as we are going to see below, objective functions for sequential decision problems come in a variety of styles.  Most important, in our view, is the need to state the objective in terms of optimizing over {\it policies}.
\end{itemize}
We suspect that the reason behind the sharp difference in styles between optimal control and Markov decision processes (adopted by reinforcement learning) is that the field of optimal control evolved from engineering, while Markov decision processes evolved out of mathematics.  The adoption of the MDP framework by reinforcement learning (which grew out of computer science and is particularly popular with less-mathematical communities) is purely historical - it was easier to make the connection between the discrete mouse-in-a-maze problem to the language of discrete Markov decision processes than stochastic optimal control.

In section \ref{sec:framework} below we are going to offer a framework that overcomes all of these limitations.  This framework, however, closely parallels the framework widely used in optimal control, with a few relatively minor modifications (and one major one).

\subsection{Bridging optimal control and reinforcement learning}
We open our discussion by noting the remarkable difference between the canonical modeling framework for optimal control, which explicitly models state variables $x_t$, controls $u_t$, information $w_t$, and transition functions, and the canonical modeling framework for reinforcement learning (inherited from Markov decision processes) which uses constructs such as state spaces, action spaces, and one-step transition matrices.  We will argue that the framework used in optimal control can be translated directly to software, whereas that used by reinforcement learning does not.

To illustrate this assertion, we note that optimal control and reinforcement learning are both addressing a sequential decision problem.  In the notation of optimal control, we would write (focusing on discrete time settings):
\bns
(x_0, u_0, w_0, x_1, u_1, w_1, \ldots, x_T).
\ens
The reinforcement learning framework, on the other hand, never models anything comparable to the information variable $w_t$.  In fact, the default setting is that we just observe the downstream state rather than modeling how we get there, but this is not universally true.

We also note that while the controls literature typically indexes variables by time, the RL community adopted the standard steady state model (see equation \eqref{eq:bellmansteadystate}) which means their variables are not indexed by time (or anything else).  Instead, they view the system as evolving in steps (or iterations).  For this reason, we are going to index variables by $n$ (as in $S^n$).

In addition, the RL community does not explicitly model an exogenous information variable.  Instead, they tend to assume when you are in a state $s$ and take action $a$, you then ``observe'' the next state $s'$.  However, any simulation in a reinforcement learning model requires creating a transition function which may (but not always) involve some random information that we call ``$w$'' (adopting, for the moment, the notation in the controls literature).  This allows us to write the sequential decision problem
\bns
(S^0, a^0, (w^1), S^1, a^1, (w^2), S^2, \ldots, S^N).
\ens
We put the $(w^n)$ in parentheses because the RL community does not explicitly model the $w^n$ process.  However, when running an RL simulation, the software will have to model this process, even if we are just observing the next state.  We use $w^{n+1}$ after taking action $a^n$ simply because it is often the case that $S^{n+1} = w^{n+1}$.

We have found that the reinforcement learning community likes to start by stating a model in terms of the MDP formal model, but then revert to the framework of stochastic control.  A good example is the presentation by \cite{lazaric2019}; slide 22 presents the MDP formal model, but when the presentation turns to present an illustration (using a simple inventory problem), it turns to the style used in the optimal control community (see slide 29).  Note that the presentation insists that the demand be stationary, which seems to be an effort to force it into the standard stationary model (see equation \eqref{eq:bellmansteadystate}). We use a much more complex inventory problem in this article, where we do not require stationarity (and which would not be required by the canonical optimal control framework).

So, we see that both optimal control and reinforcement learning are solving sequential decision problems, also known as Markov decision problems.  Sequential decision problems (decision, information, decision, information, $\ldots$) span a truly vast range of applications, as noted in section \ref{sec:communities}.  We suspect that this space is much broader than has been traditionally viewed within either of these two communities.  This is not to say that all these problems can be solved with $Q$-learning or even any Bellman-based method, but below we will identify four classes of policies that span any approach that {\it might} be used for any sequential decision problems.



The optimal control literature has its origins in problems with continuous states and actions, although the mathematical model does not impose any restrictions beyond the basic structure of sequential decisions and information (for stochastic control problems).  While optimal control is best known for the theory surrounding the structure of linear-quadratic regulation which produces the linear policy in \eqref{eq:linearcontrol}, it should not be surprising that the controls community branched into more general problems, requiring different solution strategies. These include:
\begin{itemize}
  \item Approximating the cost-to-go function $J_t(x_t)$.
  \item Determining a decision now by optimizing over a horizon $t, \ldots, t+H$ using a presumably-known model of the system (which is not always available).  This approach became known as {\it model predictive control}
  \item Specifying a parametric control law, which is typically linear in the parameters (following the style of \eqref{eq:linearcontrol}).
\end{itemize}

At the same time, the reinforcement learning community found that the performance of $Q$-learning (that is, equations \eqref{eq:qlearning1}-\eqref{eq:qlearning2}), despite the hype, did not match early hopes and expectations.  In fact, just as the optimal controls community evolved different solution methods, the reinforcement learning community followed a similar path (the same statement can be made of a number of fields in stochastic optimization).  This evolution is nicely documented by comparing the first edition of Sutton and Barto's {\it Reinforcement Learning} \citep{Sutton1998}, which focuses exclusively on $Q$-learning, with the second edition \citep{Sutton2018}, which covers methods such as Monte Carlo tree search, upper confidence bounding, and the policy gradient method.

We are going to next present (in section \ref{sec:framework}) a universal framework which is illustrated in section \ref{sec:energystorage} on a series of problems in energy storage.  Section \ref{sec:policies} will then present four classes of policies that cover every method that has been proposed in the literature, which span all the variations currently in use in both the controls literature as well as the growing literature on reinforcement learning.  We then return to the energy storage problems in section \ref{sec:policiesforenergystorage} and illustrate all four classes of policies (including a hybrid).

\section{The universal modeling framework}
\label{sec:framework}

We are going to present a universal modeling framework that covers all of the disciplines and application domains listed in section \ref{sec:communities}. The framework will end up posing an optimization problem that involves searching over {\it policies}, which are functions for making decisions.  We will illustrate the framework on a simple inventory problem using the setting of controlling battery storage (a classical stochastic control problem).

In section \ref{sec:energystorage} we will illustrate some key concepts by extending our energy storage application, focusing primarily on modeling state variables.  Then, section \ref{sec:policies} describes a general strategy for designing policies, which we are going to claim covers {\it every} solution approach proposed in the research literature (or used in practice).  Thus, we will have a path to finding solutions to {\it any} problem (but these are rarely optimal).

Before starting, we make a few notes on notation:
\begin{itemize}
  \item The controls community uses $x_t$ for state, while the reinforcement learning community adopted the widely used notation $S_t$ for state.  We have used $S_t$ partly because of the mnemonics (making it easier to remember), but largely because $x_t$ conflicts with the notation for decisions adopted by the field of math programming, which is widely used.
  \item There are three standard notational systems for decisions: $a$ for action (typically discrete), $u$ for control (typically a low-dimensional, continuous vector), and $x$, which is the notation used by the entire math programming community, where $x$ can be continuous or discrete, scalar or vector.  We adopt $x$ because of how widely it is used in math programming, and because it has been used in virtually every setting (binary, discrete, continuous, scalar or vector).  It has also been adopted in the multi-armed bandit community in computer science.
  \item The controls community uses $w_t$ which is (sadly) random at time $t$, whereas all other variables are known at time $t$.  We prefer the style that every variable indexed by time $t$ (or iteration $n$) is known at time $t$ (or iteration $n$).  For this reason, we use $W_t$ for the exogenous information that first becomes known between $t-1$ and $t$, which means it is known at time $t$. (Similarly, $W^n$ would be information that becomes known between iterations/observations $n-1$ and $n$.)
\end{itemize}

\subsection{Dimensions of a sequential decision model}
There are five elements to any sequential decision problem: state variables, decision variables, exogenous information variables, transition function, and objective function.  We briefly describe each below, returning in section \ref{sec:statevariables} to discuss state variables in more depth.  The description below is adapted from \cite{PowellEJOR2019}.

\begin{description}
  \item[State variables] - The state $S_t$ of the system at time $t$ (we might say $S^n$ after $n$ iterations) is a function of history which contains all the information that is necessary and sufficient to compute costs/rewards, constraints, and any information needed by the transition function.  The state $S_t$ typically consists of a number of dimensions which we might write as $S_t = (S_{t1}, \ldots, S_{tK})$.  This will be more meaningful when we illustrate it with an example below.

      We distinguish between the initial state $S_0$ and the dynamic state $S_t$ for $t > 0$. The initial state contains all deterministic parameters, initial values of any dynamic parameters, and initial beliefs about unknown parameters in the form of the parameters of probability distributions.  The dynamic state $S_t$ contains only information that is evolving over time.

      In section \ref{sec:statevariables}, we will distinguish different classes of state variables, including physical state variables $R_t$ (which might describe inventories or the location of a vehicle), other information $I_t$ (which might capture prices, weather, or the humidity in a laboratory), and beliefs $B_t$ (which includes the parameters of probability distributions describing unobservable parameters).  It is sometimes helpful to recognize that $(R_t,I_t)$ capture everything that can be observed perfectly, while $B_t$ represents distributions of anything that is uncertain.
%
%
  \item[Decision variables] - We use $x_t$ for decisions, where $x_t$ may be binary (e.g. for a stopping problem), discrete (e.g. an element of a finite set),  continuous (scalar or vector), integer vectors, and categorical (e.g. the attributes of a patient).  In some applications $x_t$ might have hundreds of thousands, or even millions, of {\it dimensions}, which makes the concept of ``action spaces'' fairly meaningless.  We note that entire fields of research are sometimes distinguished by the nature of the decision variable.

      We assume that decisions are made with a policy, which we might denote $X^\pi(S_t)$.  We also assume that a decision $x_t = X^\pi(S_t)$ is feasible at time $t$. We let ``$\pi$'' carry the information about the type of function $f\in\Fcal$ (for example, a linear model with specific explanatory variables, or a particular nonlinear model), and any tunable parameters $\theta \in \Theta^f$.
  \item[Exogenous information] - We let $W_t$ be any new information that first becomes known at time $t$ (that is, between $t-1$ and $t$).  This means any variable indexed by $t$ is known at time $t$.  When modeling specific variables, we use ``hats'' to indicate exogenous information.  Thus, $\Dhat_t$ could be the demand that arose between $t-1$ and $t$, or we could let $\phat_t$ be the change in the price between $t-1$ and $t$.  The exogenous information process may be stationary or nonstationary, purely exogenous or state (and possibly action) dependent.

      As with decisions, the exogenous information $W_t$ might be scalar, or it could have thousands to millions of {\it dimensions} (imagine the number of new customer requests for trips from zone $i$ to zone $j$ in an area that has 20,000 zones).

      The distribution of $W_{t+1}$ (given we are at time $t$) may be described by a known mathematical model, or we may depend on observations from an exogenous source (this is known as ``data driven'').  The exogenous information may depend on the current state and/or action, so we might write it as $W_{t+1}(S_t,x_t)$.  We will suppress this notation moving forward, but with the understanding that we allow this behavior.
  \item[Transition function] - We denote the transition function by
      \bn
      S_{t+1} = S^M(S_t,x_t,W_{t+1}), \label{eq:transition}
      \en
      where $S^M(\cdot)$ is also known by names such as system model, state equation, plant model, plant equation and transfer function.  We have chosen not to use the standard notation $f(s,x,w)$ used universally by the controls community simply because the letter $f$ is also widely used for ``functions'' in many settings.  The alphabet is very limited and the letter $f$ occupies a valuable piece of real-estate.

      An important problem class in both optimal control and reinforcement learning arises when the transition function is unknown.  This is sometimes referred to as ``model-free dynamic programming.'' There are some classes of policies that do not need a transition function, both others do, introducing the dimension of trying to learn the transition function.
  \item[Objective functions] - There are a number of ways to write objective functions in sequential decision problems.  Our default notation is to let
      \bns
      C_t(S_t,x_t) = \textwrap{the contribution of taking action $x_t$ given the information in state $S_t$.}
      \ens
      For now we are going to use the most common form of an objective function used in both dynamic programming (which includes reinforcement learning) and stochastic control, which is to maximize the expected sum of contributions:
      \bn
      \max_\pi \E_{S_0} \E_{W_1, \ldots, W_T|S_0} \left\{\sum_{t=0}^T C_t(S_t,X^\pi_t(S_t))|S_0\right\}, \label{eq:basemodel}
      \en
      where
      \bn
      S_{t+1} = S^M(S_t,X^\pi_t(S_t),W_{t+1}), \label{eq:basetransition}
      \en
      and where we are given a source of the exogenous information process
      \bn
      (S_0, W_1, W_2, \ldots, W_T).  \label{eq:baseinformation}
      \en
      We refer to equation \eqref{eq:basemodel} along with the state transition function \eqref{eq:basetransition} and exogenous information \eqref{eq:baseinformation} as the {\it base model}.  We revisit objective functions in section \ref{sec:objectivefunction}.
\end{description}

An important feature of our modeling framework is that we introduce the concept of a policy $X^\pi(S_t)$ when we describe decisions, and we search over policies in the objective function in equation \eqref{eq:basemodel}, but we do not at this point specify what the policies might look like.  Searching over policies is precisely what is meant by insisting that the control $u_t$ in equation \eqref{eq:canonicalstochasticcontrol} be ``$\Fcal_t$-measurable.''  In section \ref{sec:policies} we are going to make this much more concrete, and does not require mastering subtle concepts such as ``measurability.''  All that is needed is the understanding that a policy depends on the state variable (measurability is guaranteed when this is the case).

In other words (and as promised), we have modeled the problem without specifying how we would solve them (that is, we have not specified how we are computing the policy).  This follows our ``{\it Model first, then solve}'' approach. Contrast this with the $Q$-learning equations \eqref{eq:qlearning1} - \eqref{eq:qlearning2} which is basically an algorithm without a model, although the RL community would insist that the model is the canonical MDP framework given in section \ref{sec:reinforcementlearning}.

\subsection{State variables}
\label{sec:statevariables}
Our experience is that there is an almost universal misunderstanding of what is meant by a ``state variable.''  Not surprisingly, interpretations of the term ``state variable'' vary between communities.  An indication of the confusion can be traced to attempts to define state variables.  For example, Bellman introduces state variables with ``we have a physical system characterized at any stage by a small set of parameters, the {\it state variables}'' \citep{Be57}.  Puterman's now classic text introduces state variables with ``At each decision epoch, the system occupies a {\it state}.'' \citep{Puterman05}[p. 18] (in both cases, the italicized text was included in the original text).  As of this writing, Wikipedia offers ``A state variable is one of the set of variables that are used to describe the mathematical ‘state’ of a dynamical system.''  Note that all three references use the word ``state'' in the definition of state variable.

It has also been our finding that most books in optimal control do, in fact, include proper definitions of a state variable (our experience is that this is the only field that does this).  They all tend to say the same thing: a state variable $x_t$ is all the information needed to model the system from time $t$ onward.

Our only complaint about the standard definition used in optimal control books is that it is vague.  The definition proposed in \cite{PowellRLSO2020} (building on the definition in \cite{Po11}) refines the basic definition with the following:

\singlespace
\begin{changemargin}{1.0cm}{1.0cm}
{\small
A {\bf state variable} is:
\begin{description}
\item[a) Policy-dependent version] A function of history that, combined with the exogenous information (and a policy), is necessary and sufficient to compute the decision function (the policy), the cost/contribution function,  and the transition function.
\item[b) Optimization version] A function of history that, combined with the exogenous information, is necessary and sufficient to compute the cost or contribution function, the constraints, and the transition function.
\end{description}
}
\end{changemargin}

\doublespacetwo

There are three types of information in $S_t$:
\begin{itemize}
\item The physical state, $R_t$, which in most (but not all) applications is the state variables that are being controlled. $R_t$ may be a scalar, or a vector with element $R_{ti}$ where $i$ could be a type of resource (e.g. a blood type) or the amount of inventory at location $i$.  Physical state variables typically appear in the constraints.  We make a point of singling out physical states because of their importance in modeling resource allocation problems, where the ``state of the system'' is often (and mistakenly) equated with the physical state.
\item Other information, $I_t$, which is any information that is known deterministically not included in $R_t$.  The information state often evolves exogenously, but may be controlled or at least influenced by decisions (e.g. selling a large number of shares may depress prices).  Other information may appear in the objective function (such as prices), and the coefficients in the constraints.
\item The belief state $B_t$, which contains distributional information about unknown parameters, where we can use frequentist or Bayesian belief models.  These may come in the following styles:
    \begin{itemize}
    \item Lookup tables - Here we have a set of discrete values $x\in\Xcal = \{x_1, \ldots, x_M\}$, and we have a belief about a function (such as $f(x) = \E F(x,W)$) for each $x\in\Xcal$.
    \item Parametric belief models - We might assume that $\E F(x,W) = f(x|\theta)$ where the function $f(x|\theta)$ is known but where $\theta$ is unknown. We would then describe $\theta$ by a probability distribution.
    \item Nonparametric belief models - These approximate a function at $x$ by smoothing local information near $x$.
    \end{itemize}
    It is important to recognize that the belief state includes the parameters of a probability distribution describing unobservable parameters of the model.  For example, $B_t$ might be the mean and covariance matrix of a multivariate normal distribution, or a vector of probabilities $p^n = (p^n_k)_{k=1}^K$ where $p^n_k = Prob[\theta = \theta_k|S^n]$.  \footnote{It is not unusual for people to overlook the need to include beliefs in the state variable.  The RL tutorial \cite{lazaric2019} does this when it presents the multiarmed bandit problem,  insisting that it does not have a state variable (see slide 49). In fact, any bandit problem is a sequential decision problem where the state variable is the belief (which can be Bayesian or frequentist). This has long been recognized by the probability community that has worked on bandit problems since the 1950's (see the seminal text \cite{Dg70}).  Bellman's equation (using belief states) was fundamental to the development of Gittins indices in \cite{GiJo74} (see \cite{gittins2011} for a nice introduction to this rich area of research).  It was the concept of Gittins indices that laid the foundation for upper confidence bounding, which is just a different form of index policy.}
\end{itemize}
We feel that a proper understanding of state variables opens up the use of the optimal control framework to span the entire set of communities and applications discussed in section \ref{sec:communities}.

\subsection{Objective functions}
\label{sec:objectivefunction}
Sequential decision problems are diverse, and this is reflected in the different types of objective functions that may be used.  Our framework is insensitive to the choice of objective function, but they all still require optimizing over policies.

We begin by making the distinction between state-independent problems, and state-dependent problems.  We let $F(x,W)$ denote a state-independent problem, where we assume that neither the objective function $F(x,W)$, nor any constraints, depends on dynamic information captured in the state variable.  We let $C(S,x)$ capture state-dependent problems, where the objective function (and/or constraints) may depend on dynamic information.

Throughout we assume problems are formulated over finite time horizons.  This is the most standard approach in optimal control, whereas the reinforcement learning community adopted the style of Markov decision processes to model problems over infinite time horizons.  We suspect that the difference reflects the history of optimal control, which is based on solving real engineering problems, and Markov decision processes, with its roots in mathematics and stylized problems.

In addition to the issue of state-dependency, we make the distinction between optimizing the cumulative reward versus the final reward.  When we combine state dependency and the issue of final vs. cumulative reward, we obtain four objective functions.  We present these in the order: 1) State-independent, final reward, 2) state-independent, cumulative reward, 3) state-dependent, cumulative reward, and 4) state-dependent, final reward (the last class is the most subtle).
\begin{description}
\item[State-independent functions] These are pure learning problems, where the {\it problem} does not depend on information in the state variable.  The only state variable is the belief about an unknown function $\E_W F(x,W)$.
    \begin{description}
      \item[1) Final reward] This is the classical stochastic search problem.  Here we go through a learning/training process to find a final design/decision $x^{\pi,N}$, where $\pi$ is our search policy (or algorithm), and $N$ is the budget.  We then have to test the performance of the policy by simulating $\What$ using
          \bn
          \max_\pi \E_{S^0} \E_{W^1, \ldots, W^N|S^0} \E_{\What|S^0} F(x^{\pi,N},\What), \label{eq:stateindependentfinalreward}
          \en
          where $x^{\pi,N}$ depends on $S^0$ and the experiments $W^1, \ldots, W^N$, and where $\What$ represents the process of testing the design $x^{\pi,N}$.
      \item[2) Cumulative reward] This describes problems where we have to learn in the field, which means that we have to optimize the sum of the rewards we earn, eliminating the need for a final testing pass.  This objective is written
          \bn
          \max_\pi \E_{S^0} \E_{W^1, \ldots, W^N|S^0}\sum_{n=0}^{N-1}F(X^\pi(S^n),W^{n+1}). \label{eq:stateindependentcumulativereward}
          \en
    \end{description}
\item[State-dependent functions] This describes the massive universe of problems where the objective and/or the constraints depend on the state variable which may or may not be controllable.
    \begin{description}
      \item[3) Cumulative reward] This is the version of the objective function that is most widely used in stochastic optimal control (as well as Markov decision processes).  We switch back to time-indexing here since these problems are often evolving over time (but not always).  We write the contribution in the form $C(S_t,x_t,W_{t+1})$ to help with the comparison to $F(x,W)$.
          \bn
          \hspace{-.2in}\max_\pi \E_{S_0} \E_{W_1, \ldots, W_T|S_0}  \left\{\sum_{t=0}^T C(S_t,X^\pi(S_t),W_{t+1})|S_0\right\}. \label{eq:statedependentcumulativereward}
          \en
      \item[4) Final reward] This is the objective function that describes optimization algorithms (represented as $\pi^{lrn}$) optimizing a time-staged, state-dependent objective.  This is the objective that should be used when finding the best algorithm for a dynamic program/stochastic control problem, yet has been almost universally overlooked as a sequential decision problem.  The objective is given by
          \bn
          \max_{\pi^{lrn}}\E_{S^0} \E^{\pi^{lrn}}_{W^1, \ldots, W^N|S^0} \E^{\pi^{imp}}_{S|S^0}\E_{\What|S^0} C(S,X^{\pi^{imp}}(S|\theta^{\pi^{imp}}),\What).  \label{eq:statedependentfinalreward}
          \en
          where $\pi^{lrn}$ is the learning policy (or algorithm), while $\pi^{imp}$ is the implementation policy that we are learning through $\pi^{lrn}$.  We note that we use the learning policy $\pi^{lrn}$ to learn the parameters $\theta^{\pi^{imp}}$ that govern the behavior of the implementation policy.
    \end{description}
\end{description}
There are many problems that require more complex objective functions such as the best (or worst) performance in a time period, across all time periods.  In these settings we cannot simply sum the contributions across time periods (or iterations).  For this purpose, we introduce the operator $\rhooperator$ which takes as input the entire sequence of contributions.  We would write our objective function as

{\small
\bn
\max_\pi \E_{S^0} \E_{W^1, \ldots, W^N|S^0}~ \rhooperator\big(C_0(S_0,X^\pi(S_0),W_{1}), C_2(S_2,X^\pi(S_2),W_{3}),\ldots,C_t(S_t,X^\pi_t(S_t),W_{t+1}), \ldots \big). \label{eq:basemodelrisk}
\en
}
The objective in \eqref{eq:basemodelrisk}, through creative use of the operator $\rhooperator$, subsumes all four objectives \eqref{eq:stateindependentfinalreward} - \eqref{eq:statedependentfinalreward}.  However, we feel that generality comes at a cost of clarity.

The controls community, while also sharing an interest in risk, is also interested in stability, an issue that is important in settings such as controlling aircraft and rockets.  While we do not address the specific issue of designing policies to handle stability, we make the case that the problem of searching over policies remains the same; all that has changed is the metric.

All of these objectives can be written in the form of {\it regret} which measures the difference between the solution we obtain and the best possible.  Regret is popular in the learning community where we compare against the solution that assumes perfect information.  A comparable strategy compares the performance of a policy against what can be achieved with perfect information about the future (widely known as a posterior bound).

\subsection{Notes}
It is useful to  list some similarities (and differences) between our modeling framework and that used in stochastic optimal control:
\begin{itemize}
  \item[1)] The optimal control framework includes all five elements, although we lay these out more explicitly.
  \item[2)] We use a richer understanding of state variables, which means that we can apply our framework to a much wider range of problems than has traditionally been considered in the optimal control literature.  In particular, {\it all} the fields and problem areas in section \ref{sec:communities} fit this framework, which means we would say that all of these are ``optimal control problems.''
  \item[3)] The stochastic control modelling framework uses $w_t$ as the information that will arrive between $t$ and $t+1$, which means it is random at time $t$.  We let $W_t$ be the information that arrives between $t-1$ and $t$, which means it is known at time $t$.  This means we write our transition as
      \bns
      S_{t+1} = S^M(S_t,x_t,W_{t+1}).
      \ens
      This notation makes it explicitly clear that $W_{t+1}$ is not known when we determine decision $x_t$.
  \item[4)] We recognize a wider range of objective functions, which expands the problem classes to offline and online applications, active learning (bandit) problems, and hybrids.
  \item[5)] We formulate the optimization problem in terms of optimizing over policies, without prejudging the classes of policies. We describe four classes of policies in section \ref{sec:policies} that we claim are universal: they cover all the strategies that have been proposed or used in practice.  This also opens the door to creating hybrids that combine two or more classes of policies.
\end{itemize}
The first four items are relatively minor, highlighting our belief that stochastic control is fundamentally the most sound of all the modeling frameworks used by any of the communities listed in section \ref{sec:communities}.  However, the fifth item is a significant transition from how sequential decision problems are approached today.

Many have found that $Q$-learning often does not work well.  In fact, $Q$-learning, as with all approximate dynamic programming algorithms, tend to work well only on a fairly small set of problems.  Our experience is that approximate dynamic programming ($Q$-learning is a form of approximate dynamic programming) tends to work well when we can exploit the structure of the value function.  For example, ADP has been very successful with some very complex, high-dimensional problems in fleet management (see \cite{Simao2009a} and \cite{Bouzaiene-Ayari2016}) where the value functions were convex.  However, vanilla approximation strategies (e.g. using simple linear models for value function approximations) can work very poorly even on small inventory problems (see \cite{JiPhPo14} for a summary of experiments which compare results against a rigorous benchmark).  Furthermore, as we will see in section \ref{sec:policies} below, there are a range of policies that do not depend on value functions that are natural choices for many applications.

\section{Energy storage illustration}
\label{sec:energystorage}

\begin{figure}[tb]
\begin{center}
    \includegraphics[width=4.0in]{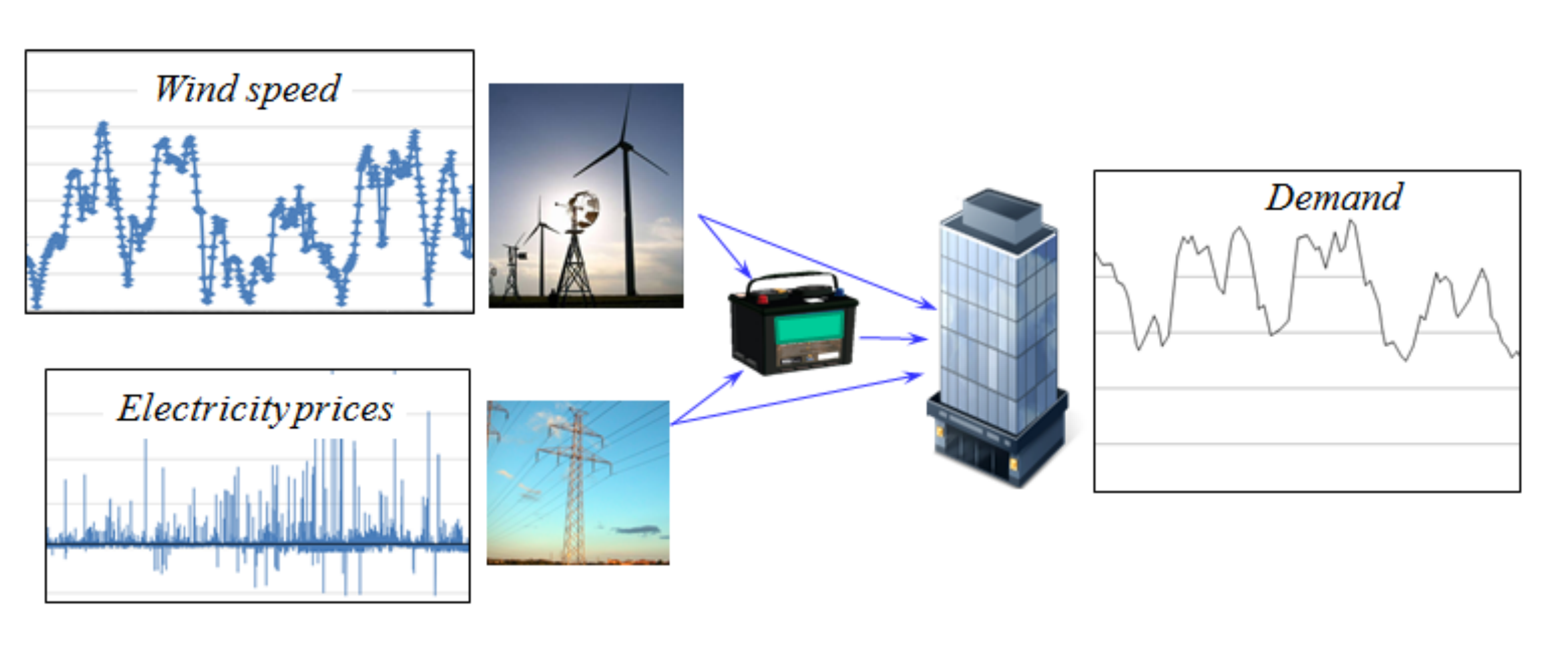}
    \caption{Energy system consisting of wind farms, grid, market, and battery storage.}
    \label{fig:energysystem}
\end{center}
\end{figure}
We are going to illustrate our modeling framework using the energy system depicted in figure \ref{fig:energysystem}, which consists of a wind farm (where energy is free but with high variability in supply), the grid (which has unlimited supply but highly stochastic prices), a market (which exhibits very time-dependent, although relatively predictable, demands), and an energy storage device (we will assume it is a battery).  While small, this rich system introduces a variety of modeling and algorithmic challenges.

We are going to demonstrate how to model this problem, starting with a simple model and then expanding to illustrate some modeling devices.  We will translate each variation into the five core components: state variables, decision variables, exogenous information variables, transition function, and objective function.

\subsection{A basic energy storage problem}


\begin{description}
  \item[State variables] State $S_t = (R_t, D_t, E_t, p_t)$ where
  \bns
  R_t &=& \textwrap{Energy in the battery at time $t$,}\\
  E_t &=& \textwrap{Power being produced by the wind farms at time $t$,}\\
  D_t &=& \textwrap{Demand for power at time $t$,}\\
  p_t &=& \textwrap{Price of energy on the grid at time $t$.}
  \ens
  Note that it is necessary to go through the rest of the model to determine which variables are needed to compute the objective function, constraints, and transition function.
  \item[Decision variables] $x_t = (x^{GB}_t, x^{GD}_t, x^{EB}_t, x^{ED}_t, x^{BD}_t)$ where
  \bns
  x^{GB}_t &=& \textwrap{Flow of energy from grid to battery ($x^{GB}_t > 0$) or back ($x^{GB}_t < 0$),}\\
  x^{GD}_t &=& \textwrap{Flow of energy from grid to demand,}\\
  x^{EB}_t &=& \textwrap{Flow of energy from wind farm to battery,}\\
  x^{ED}_t &=& \textwrap{Flow of energy from wind farm to demand,}\\
  x^{BD}_t &=& \textwrap{Flow of energy from battery to demand.}
  \ens
  These decision variables have to obey the constraints:
  \bn
  x^{EB}_t + x^{ED} & \leq & E_t, \label{eq:energyconstraint1}\\
  x^{GD}_t + x^{BD}_t + x^{ED}_t & = & D_t,\label{eq:energyconstraint2}\\
  x^{BD}_t & \leq & R_t,\label{eq:energyconstraint3}\\
  x^{GD}_t, x^{EB}_t, x^{ED}_t, x^{BD}_t & \geq & 0.\label{eq:energyconstraint4}
  \en
  Finally, we introduce the policy (function) $X^\pi(S_t)$ that will return a feasible vector $x_t$.  We defer to later the challenge of designing a good policy.
  \item[Exogenous information variables] $W_{t+1} = (\Ehat_{t+1}, \Dhat_{t+1}, \phat_{t+1})$, where
  \bns
  \Ehat_{t+1} &=& \textwrap{The change in the power from the wind farm between $t$ and $t+1$,}\\
  \Dhat_{t+1} &=& \textwrap{The change in the demand between $t$ and $t+1$,}\\
  \phat_{t+1} &=& \textwrap{The price charged at time $t+1$ as reported by the grid.}
  \ens
  We note that the first two exogenous information variables are defined as changes in values, while the last (price) is reported directly from an exogenous source.
  \item[Transition function] $S_t = S^M(S_t,x_t,W_{t+1})$:
  \bn
  R_{t+1} &=& R_t + \eta (x^{GB}_t + x^{EB}_t - x^{BD}_t),\label{eq:energytransitionR} \\
  E_{t+1} &=& E_t + \Ehat_{t+1},\label{eq:energytransitionE}\\
  D_{t+1} &=& D_t + \Dhat_{t+1},\label{eq:energytransitionD}\\
  p_{t+1} &=& \phat_{t+1}.\label{eq:energytransitionp}
  \en
  Note that we have illustrated here a controllable transition \eqref{eq:energytransitionR}, two where the exogenous information is represented as the change in a process (equations \eqref{eq:energytransitionE} and \eqref{eq:energytransitionD}), and one where we directly observe the updated price \eqref{eq:energytransitionp}.  This means that the processes $E_t$ and $D_t$ are first-order Markov chains (assuming that $\Ehat_t$ and $\Dhat_t$ are independent across time), while the price process would be described as ``model free'' or ``data driven'' since we are not assuming that we have a mathematical model of the price process.
  \item[Objective function] We wish to find a policy $X^\pi(S_t)$ that solves
  \bns
  \max_\pi \E_{S_0} \E_{W_1,\ldots,W_T|S_0} \left\{\sum_{t=0}^T C(S_t, X^\pi(S_t))|S_0\right\},
  \ens
  where $S_{t+1} = S^M(S_t,x_t=X^\pi(S_t),W_{t+1})$ and where we are given an information process
  \bns
  (S_0, W_1, W_2, \ldots, W_T).
  \ens
\end{description}
Normally, we would transition at this point to describe how we are modeling the uncertainty in the information process $(S_0, W_1, W_2, \ldots, W_T)$, and then describe how to design policies.  For compactness, we are going to skip these steps now, and instead illustrate how to model a few problem variations that can often cause confusion.

\subsection{With a time-series price model}
We are now going to make a single change to the model above.  Instead of assuming that prices $p_t$ are provided exogenously, we are going to assume we can model them using a time series model given by
\bn
p_{t+1} &=& \theta_0 p_t + \theta_1 p_{t-1} + \theta_2 p_{t-2} + \varepsilon_{t+1}. \label{eq:priceprocesswithoutlearning}
\en
A common mistake is to say that $p_t$ is the ``state'' of the price process, and then observe that it is no longer Markovian (it would be called ``history dependent''), but ``it can be made Markovian by expanding the state variable,'' which would be done by including $p_{t-1}$ and $p_{t-2}$ (see \cite{cinlar2011} for an example of this).  According to our definition of a state variable, the state is all the information needed to model the process from time $t$ onward, which means that the state of our price process is $(p_t, p_{t-1}, p_{t-2})$.  This means our system state variable is now
\bns
S_t = \big((R_t, D_t, E_t), (p_t,p_{t-1},p_{t-2})\big)).
\ens
We then have to modify our transition function so that the ``price state variable'' at time $t+1$ becomes $(p_{t+1}, p_t, p_{t-1})$.

\subsection{With passive learning}
We implicitly assumed that our price process in equation \eqref{eq:priceprocesswithoutlearning} was governed by a model where the coefficients $\theta = (\theta_0, \theta_1, \theta_2)$ were known.  Now assume that the price $p_{t+1}$ depends on prices over the last three time periods, which means we would write
\bn
p_{t+1} &=& \thetabar_{t0} p_t + \thetabar_{t1} p_{t-1} + \thetabar_{t2} p_{t-2} + \varepsilon_{t+1}. \label{eq:priceprocesspassivelearning}
\en
Here, we have to adaptively update our estimate $\thetabar_t$ which we can do using recursive least squares.  To do this, let
\bns
\pbar_t                      &=& (p_t, p_{t-2}, p_{t-2})^T,\\
\Fbar_t(\pbar_t|\thetabar_t) &=& (\pbar_t)^T \thetabar_t
\ens
We perform the updating using a standard set of updating equations given by
\bn
\thetabar_{t+1} &=& \thetabar_t + \frac{1}{\gamma_t} M_t \pbar_t \varepsilon_{t+1}, \label{eq:learning1} \\
\varepsilon_{t+1} &=& \Fbar_t(\pbar_t|\thetabar_t) - p_{t+1}, \label{eq:learning2} \\
M_{t+1} &=& M_t - \frac{1}{\gamma_t} M_t (\pbar_t) (\pbar_t)^T M_t, \label{eq:learning3} \\
\gamma_t &=& 1- (\pbar_t)^T M_t \pbar_t. \label{eq:learning4}
\en
To compute these equations, we need the three-element vector $\thetabar_t$ and the $3 \times 3$ matrix $M_t$.  These then need to be added to our state variable, giving us
\bns
S_t = \big((R_t, D_t, E_t), (p_t,p_{t-1},p_{t-2}), (\thetabar_t,M_t)\big).
\ens
We then have to include equations \eqref{eq:learning1} - \eqref{eq:learning4} in our transition function.

\subsection{With active learning}
We can further generalize our model by assuming that our decision $x^{GB}_t$ to buy or sell energy from or to the grid can have an impact on prices.  We might propose a modified price model given by
\bn
p_{t+1} &=& \thetabar_{t0} p_t + \thetabar_{t1} p_{t-1} + \thetabar_{t2} p_{t-2} + \thetabar_{t3} x^{GB}_t + \varepsilon_{t+1}. \label{eq:priceprocessactivelearning}
\en
All we have done is introduce a single term $\thetabar_{t3} x^{GB}_t$ to our price model.  Assuming that $\theta_3 >0$, this model implies that purchasing power from the grid ($x^{GB}_t > 0$) will increase grid prices, while selling power back to the grid ($x^{GB}_t <0$) decreases prices.  This means that purchasing a lot of power from the grid (for example) means we are more likely to observe higher prices, which may assist the process of learning $\theta$.  When decisions control or influence what we observe, then this is an example of {\it active learning}.

This change in our price model does not affect the state variable from the previous model, aside from adding one more element to $\thetabar_t$, with the required changes to the matrix $M_t$.  The change will, however, have an impact on the policy.  It is easier to learn $\theta_{t3}$ by varying $x^{GB}_t$ over a wide range, which means trying values of $x^{GB}_t$ that do not appear to be optimal given our current estimate of the vector $\thetabar_t$.  Making decisions partly just to learn (to make better decisions in the future) is the essence of {\it active learning}, best known in the field of multiarmed bandit problems.

\subsection{With rolling forecasts}
Forecasting is such a routine activity in operational problems, it may come as a surprise that we have been modelling these problems incorrectly.

Assume we have a forecast $f^E_{t,t+1}$ of the energy $E_{t+1}$ from wind, which means
\bn
E_{t+1} = f^E_{t,t+1} + \varepsilon_{t+1,1},  \label{eq:energytransitionE1}
\en
where $\varepsilon_{t+1,1} \sim N(0, \sigma^2_\varepsilon)$ is the random variable capturing the one-period-ahead error in the forecast.

Equation \eqref{eq:energytransitionE1} effectively replaces equation \eqref{eq:energytransitionE} in the transition function for the base model.  However, it introduces a new variable, the forecast $f^E_{t,t+1}$, which must now be added to the state variable.  This means we now need a transition equation to describe how $f^E_{t,t+1}$ evolves over time.  We do this by using a two-period-ahead forecast, $f^E_{t,t+2}$, which is basically a forecast of $f^E_{t+1,t+2}$, plus an error, giving us
\bn
f^E_{t+1,t+2} = f^E_{t,t+2} + \varepsilon_{t+1,2},  \label{eq:energytransitionE2}
\en
where $\varepsilon_{t+1,2} \sim N(0, 2\sigma^2_\varepsilon)$ is the two-period-ahead error (we are assuming that the variance in a forecast increases linearly with time).  Now we have to put $f^E_{t,t+2}$ in the state variable, which generates a new transition equation.   This generalizes to
\bn
f^E_{t+1,t'} = f^E_{t,t'} + \varepsilon_{t+1,t'-t},  \label{eq:energytransitionEt}
\en
where $\varepsilon_{t+1,t'-t} \sim N(0, (t'-t)\sigma^2_\varepsilon)$.

This stops, of course, when we hit the planning horizon $H$.  This means that we now have to add
\bns
f^E_t = (f^E_{tt'})_{t'=t+1}^{t+H}
\ens
to the state variable, with the transition equations \eqref{eq:energytransitionEt} for $t'=t+1, \ldots, t+H$.  Combined with the learning statistics, our state variable is now
\bns
S_t = \big((R_t, D_t, E_t), (p_t,p_{t-1},p_{t-2}), (\thetabar_t,M_t),f^E_t\big).
\ens

It is useful to note that we have a nice illustration of the three elements of our state variable:
\bns
(R_t, D_t, E_t) &=& \textwrap{The physical state variables (note that they all appear in the right hand side of constraints \eqref{eq:energyconstraint1}-\eqref{eq:energyconstraint4}),}\\
(p_t, p_{t-1}, p_{t-2}) &=& \textwrap{other information,}\\
((\thetabar_t,M_t),f^E_t) &=& \textwrap{the belief state, since these parameters determine the distribution of belief about variables that are not known perfectly.}
\ens

\subsection{Remarks}
We note that all the models illustrated in this section are sequential decision problems, which means that all of them can be described as either stochastic control problems, or reinforcement learning problems.  This is true whether state variables or decision/control variables are scalar or vector, discrete or continuous (or mixed).  We have, however, assume that time is either discrete or discretized.

Energy storage is a form of inventory problem, which is the original stochastic control problem used by Bellman to motivate his work on dynamic programming \citep{BeGlGr55}, and is even used today by the reinforcement learning community \citep{lazaric2019}.  However, we have never seen the variations that we illustrated here solved by any of these communities.

In section \ref{sec:policies} we are going to present four classes of policies, and then illustrate, in section \ref{sec:policiesforenergystorage}, that each of the four classes (including a hybrid) can be applied to the full range of these energy storage problems.  We are then going to show that both communities (optimal control and reinforcement learning) use methods that are drawn from each of the four classes, but apparently without an awareness that these are instances in broader classes, that can be used to solve complex problems.


\section{Designing policies}
\label{sec:policies}
There are two fundamental strategies for creating policies:
\begin{description}
  \item[Policy search] - Here we use any of the objective functions  \eqref{eq:stateindependentfinalreward} - \eqref{eq:basemodelrisk} to search within a family of functions to find the policy that works best.  This means we have to a) find a class of function and b) tune any parameters.  The challenge is finding the right family, and then performing the tuning (which can be hard).
  \item[Lookahead approximations] - Alternatively, we can construct policies by approximating the impact of a decision now on the future.  The challenge here is designing and computing the approximation of the future (this is also hard).
\end{description}
Either of these approaches can yield optimal policies, although in practice this is rare.  Below we show that each of these strategies can be further divided into two classes, creating four (meta)classes of policies for making decisions.  We make the claim that these are universal, which is to say that {\it any} solution approach to any sequential decision problem will use a policy drawn from one of these four classes, or a hybrid of two or more classes.


\subsection{Policy search}
Policy search involves tuning and comparing policies using the objective functions \eqref{eq:stateindependentfinalreward} - \eqref{eq:basemodelrisk} so that they behave well when averaged over a set of sample paths.  Assume that we have a class of functions $\Fcal$, where for each function $f\in\Fcal$, there is a parameter vector $\theta\in\Theta^f$ that controls its behavior.  Let $X^\pi(S_t|\theta)$ be a function in class $f\in\Fcal$ parameterized by $\theta\in\Theta^f$, where $\pi = (f,\theta), ~f\in\Fcal, \theta\in\Theta^f$.  Policy search involves finding the best policy using
\bn
\max_{\pi\in(\Fcal,\Theta^f)} \E_{S_0} \E_{W_1,\ldots,W_T|S_0} \left\{\sum_{t=0}^T C(S_t, X^\pi(S_t))|S_0\right\}. \label{eq:policysearchobjectiveexpectation}
\en
In special cases, this can produce an optimal policy, as we saw for the case of linear-quadratic regulation (see equation \eqref{eq:linearcontrol}).

Since we can rarely find optimal policies using \eqref{eq:policysearchobjectiveexpectation}, we have identified two sub-classes within the policy search class:
\begin{description}
   \item[Policy function approximations (PFAs)] - Policy function approximations can be lookup tables, parametric or nonparametric functions, but the most common are parametric functions.  This could be a linear function such as
       \bns
       X^\pi(S_t|\theta) = \theta_0 + \theta_1 \phi_1(S_t) + \theta_2 \phi_2(S_t) + \ldots,
       \ens
       which parallels the linear control law in equation \eqref{eq:linearcontrol} (these are also known as ``affine policies'').  We might also use a nonlinear function such as an order-up-to inventory policy, a logistics curve, or a neural network. Typically there is no guarantee that a PFA is in the optimal class of policies.  Instead, we search for the best performance within a class.
   \item[Cost function approximations (CFAs)] - A CFA is
        \bns
        X^\pi(S_t|\theta) = \argmax_{x\in\Xcal^\pi_t(\theta)} \Cbar^\pi_t(S_t,x|\theta),
        \ens
        where $\Cbar^\pi_t(S_t,x|\theta)$ is a parametrically modified cost function, subject to a parametrically modified set of constraints. A popular example known to the computer science community is interval estimation where a discrete alternative $x\in \Xcal = \{x_1, \ldots, x_M\} $ is chosen which maximizes
        \bns
        X^{IE}(S^n|\theta^{IE}) = \argmax_{x\in\Xcal}\big(\mubar^n_x + \theta^{IE} \sigmabar^n_x\big)
        \ens
        where $\mubar^n_x$ is the current estimate of $\E_W F(x,W)$ after $n$ experiments, and where $\sigmabar^n_x$ is the standard deviation of the statistic $\mubar^n_x$.  Here, $\theta^{IE}$ is a parameter that has to be tuned.

        CFAs are widely used for solving large scale problems such as scheduling an airline or planning a supply chain. For example, we might introduce slack into a scheduling problem, or buffer stocks for an inventory problem.
\end{description}
Policy search is best suited when the policy has clear structure, such as inserting slack in an airline schedule, or selling a stock when the price goes over some limit.  Neural networks have become popular recently because they assume no structure, but the price of this generality is that extensive tuning is needed.  We urge caution with the use of high-dimensional architectures such as neural networks.  There are many problems where we expect the policy to exhibit structure, such as increasing the dosage of a drug with the weight of a patient, or setting the bid price of a stock as a function of market indicators.  Neural networks do not offer these guarantees, and would require a tremendous amount of training to produce this behavior.

\subsection{Lookahead approximations}
\label{sec:lookaheadapproximations}
\begin{figure}[tb]
\begin{center}
    \includegraphics[width=4.0in]{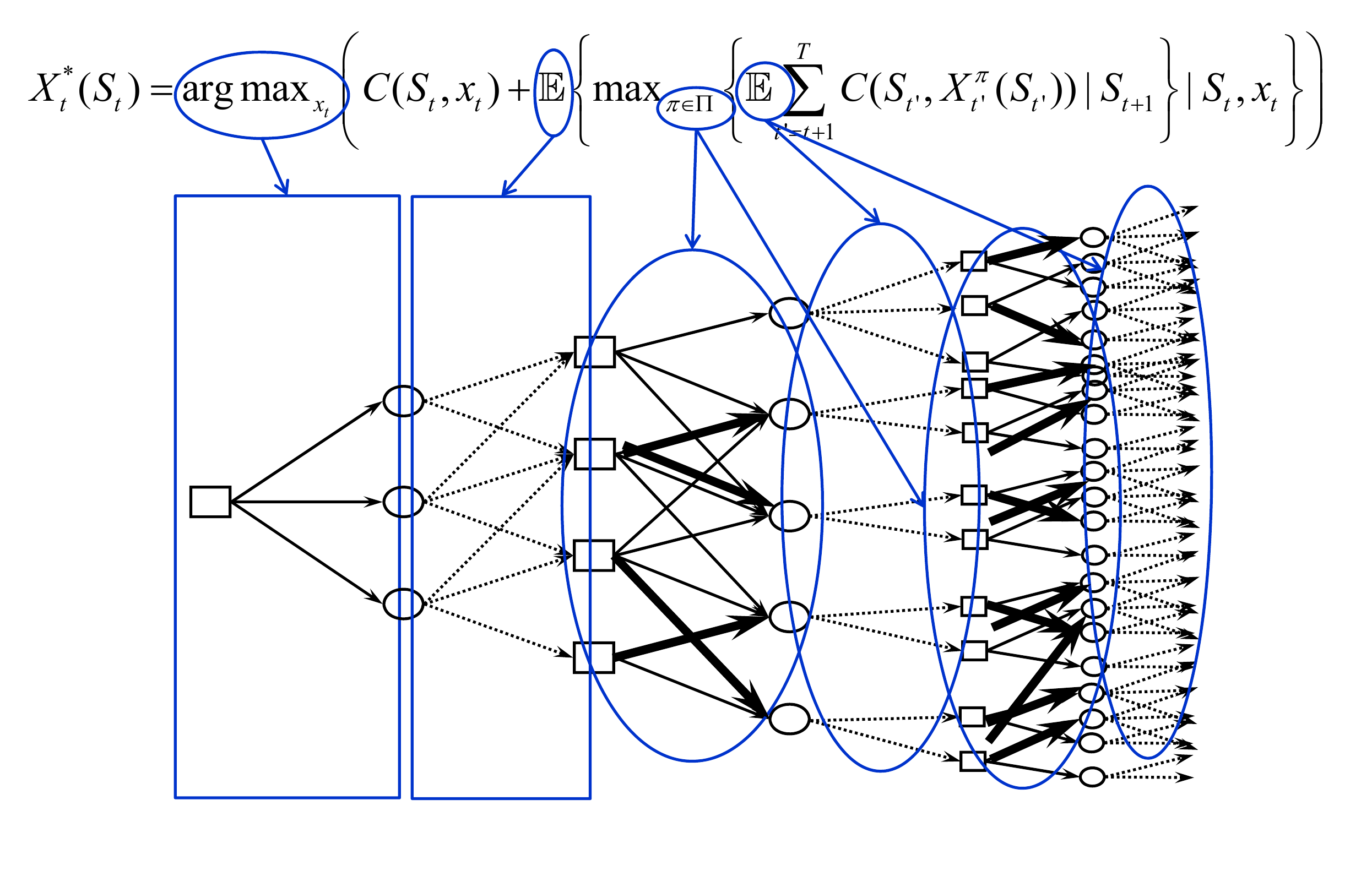}
    \caption{Relationship between the stochastic lookahead policy and a decision tree, showing initial decision, initial expectation, and then the decisions made for each state in the future (which is the lookahead policy ${\tilde \pi}$).}
    \label{fig:decisiontree}
\end{center}
\end{figure}

Just as we can, in theory, find an optimal policy using policy search, we can also find an optimal policy by modeling the downstream impact of a decision made now on the future. This can be written
\bn
X^*_t(S_t) = \argmax_{x_t} \hspace{-.03in}\left(\hspace{-.03in}C(S_t,x_t) + \E \left\{ \left. \max_\pi  \E \left\{\left.\sum_{t'=t+1}^T C(S_{t'},X^\pi_{t'}(S_{t'})) \right| S_{t+1}\right\}\right|S_t,x_t\right\}\right)\hspace{-.05in}. \label{eq:optimalLApolicy}
\en
Equation \eqref{eq:optimalLApolicy} is daunting, but can be parsed in the context of a decision tree with discrete actions and discrete random outcomes (see figure \ref{fig:decisiontree}).  The states $S_{t'}$ correspond to the square nodes in the decision tree.  The state $S_t$ is the initial node, and the actions $x_t$ are the initial actions.  The first expectation is over the first set of random outcomes $W_{t+1}$ (out of the outcome nodes resulting from each decision $x_t$). The imbedded policy $\pi$ is the choice of decision for {\it each} decision node (state) over the horizon $t+1, \ldots, T$.  The second expectation is over $W_{t+2}, \ldots, W_T$.

In practice, a stochastic lookahead policy is generally impossible to compute (decision trees grow exponentially).  There are two broad strategies for approximating the lookahead model:
\begin{description}
   \item[Value function approximations (VFAs)] - Our first approach is to replace the entire term capturing the future in \eqref{eq:optimalLApolicy} with an approximation of the value function (the controls community uses the term cost-to-go function).  We can do this in two ways.  The first is to replace the function starting at $S_{t+1}$ with a value function $V_{t+1}(S_{t+1})$ giving us
       \bn
       X^{VFA}_t(S_t) = \argmax_{x_t} \left(C(S_t,x_t) + \E \left\{\Vbar_{t+1}(S_{t+1})|S_t\right\}\right) \label{eq:vfapredecision}
       \en
       where $S_{t+1} = S^M(S_t,x_t,W_{t+1})$, and where the expectation is over $W_{t+1}$ conditioned on $S_t$ (some write the conditioning as dependent on $S_t$ and $x_t$).  Since we generally cannot compute $V_{t+1}(S_{t+1})$, we can use various machine learning methods to replace it with some sort of approximation $\Vbar_{t+1}(S_{t+1})$ called the value function approximation.

       The second way is to approximate the function around the post-decision state $S^x_t$ (this is the state immediately after a decision is made), which eliminates the expectation \eqref{eq:vfapredecision}, giving us
       \bn
       X^{VFA}_t(S_t) = \argmax_{x_t} \left(C(S_t,x_t) + \Vbar^x_t(S^x_t)\right). \label{eq:vfapostdecision}
       \en
       The benefit of using the post-decision value function approximation is that it eliminates the expectation from within the max operator.  This has proven to be especially useful for problems where $x_t$ is a vector, and $V^x_t(S^x_t)$ is a convex function of $S^x_t$.

       There is by now an extensive literature on the use of value function approximations that have evolved under names such as heuristic dynamic programming \citep{SiBaPo04}, neuro-dynamic programming \citep{BeTs96}, adaptive dynamic programming (\cite{Murray2002}, \cite{LewisVrabie2009}), approximate dynamic programming \citep{Po11}, and reinforcement learning \citep{Sutton1998}.  While the use of value function approximations has tremendous appeal, it is no panacea.  Our experience is that this approach works best only when we can exploit problem structure.
   \item[Direct lookahead (DLAs)] There are many problems where it is just not possible to compute sufficiently accurate VFAs.  When all else fails, we have to resort to a direct lookahead, where we replace the lookahead expectation and optimization in \eqref{eq:optimalLApolicy} with an {\it approximate lookahead model}.

       The most widely used approximation strategy is to use a deterministic lookahead, often associated with model predictive control, although it is more accurate to refer to any policy based on solving a lookahead model as model predictive control.  We can create an approximate (stochastic) lookahead model that we represent as the following sequence
       \bns
       (S_t,x_t, \Wtilde_{t,t+1}, \Stilde_{t,t+1}, \xtilde_{t,t+1}, \ldots, \Wtilde_{tt'}, \Stilde_{tt'}, \xtilde_{tt'}, \Wtilde_{t,t'+1}, \ldots).
       \ens
       We use tilde-variables to indicate variables within the lookahead model.  Each tilde-variable is indexed by $t$ (the time at which the lookahead model is being formed) and $t'$ (the time within the lookahead model). Our lookahead policy might then be written

       {\small
       \bn
       X^{DLA}_t(S_t) \hspace{-0.10in}&=&\hspace{-0.10in} \argmax_{x_t} \left(C(S_t,x_t) + \Etilde \left\{\max_{{\tilde \pi}} \Etilde \left\{\sum_{t'=t+1}^T C(\Stilde_{tt'},\Xtilde^{\tilde \pi}(\Stilde_{tt'})) | \Stilde_{t,t+1}\right\}|S_t,x_t\right\}\right). \label{eq:optimalpolicyLAapproxpi}
       \en
       }

       Typically the approximate expectations $\Etilde$ are computed using Monte Carlo sampling, although we can use a deterministic forecast.  The real challenge is the lookahead policy $\Xtilde^{\tilde \pi}(\Stilde_{tt'})$ which may take any form.  This policy is also known as a ``rollout policy'' where it is used in combinatorial optimization \citep{Bertsekas1997}, and Monte Carlo tree search (\cite{Chang2005a}, \cite{coulom2007}, \cite{Browne2012}).

       One possibility for the lookahead policy is to use a simpler parameterized policy that we might write $\Xtilde^{\tilde \pi}(\Stilde_{tt'}|{\tilde \theta})$.  In this case, the $\max_{{\tilde \pi}}$ operator would be replaced with $\max_{\tilde \theta}$, but even this simpler problem means that we are finding the best parameter ${\tilde \theta}$ for each state $\Stilde_{t,t+1}$, which means we are really looking for a function ${\tilde \theta}(s)$ where $s=\Stilde_{t,t+1}$.  A simpler alternative would be to fix a single parameter $\theta$ which means we now have a parameterized lookahead policy given by

       {\small
       \bn
       X^{DLA}_t(S_t|\theta) \hspace{-0.10in}&=&\hspace{-0.10in} \argmax_{x_t} \left(C(S_t,x_t) + \Etilde \left\{ \Etilde \left\{\sum_{t'=t+1}^T C(\Stilde_{tt'},\Xtilde^{\tilde \pi}(\Stilde_{tt'}|\theta)) | \Stilde_{t,t+1}\right\}|S_t,x_t\right\}\right). \label{eq:optimalpolicyLAapproxpi}
       \en
       }
       This version no longer has the imbedded $\max_{{\tilde \theta}}$, but we still have to tune $\theta$ in the policy $X^{DLA}_t(S_t|\theta)$.

       Another strategy for computing a stochastic lookahead policy is to use Monte Carlo tree search, a term coined by \citep{coulom2007} but first proposed in \citep{Chang2005a} (see \cite{Browne2012} for a tutorial in the context of deterministic problems).  This strategy searches forward in time, using methods to limit the full enumeration of the tree.  Monte Carlo tree search gained prominence from the role it played in creating AlphaGo for playing the Chinese game of Go, which was the first system to beat world class Go players (see \cite{Fu2017} for a nice review of the history of MCTS and AlphaGo).

       It is important to emphasize that designing a policy using a stochastic lookahead (even a simplified stochastic lookahead) means solving a stochastic optimization problem within the policy.  Recall that our stochastic optimization problem is the base model given by any of the objective functions described earlier (equations \eqref{eq:stateindependentfinalreward} - \eqref{eq:basemodelrisk}). Equation \eqref{eq:optimalpolicyLAapproxpi} represents the simplified stochastic optimization problem, which has to be solved at each time period.

       Figure \ref{fig:rollingDLA} depicts the process of simulating a direct lookahead (the figure uses a deterministic lookahead, but the same process would be used with any direct lookahead).  This is what is needed to do any parameter tuning for the DLA.  Not surprisingly, stochastic lookaheads can be computationally difficult to solve, which makes it particularly difficult to run simulations to do parameter tuning.
\end{description}
\begin{figure}[tb]
\begin{center}
    \includegraphics[width=4.0in]{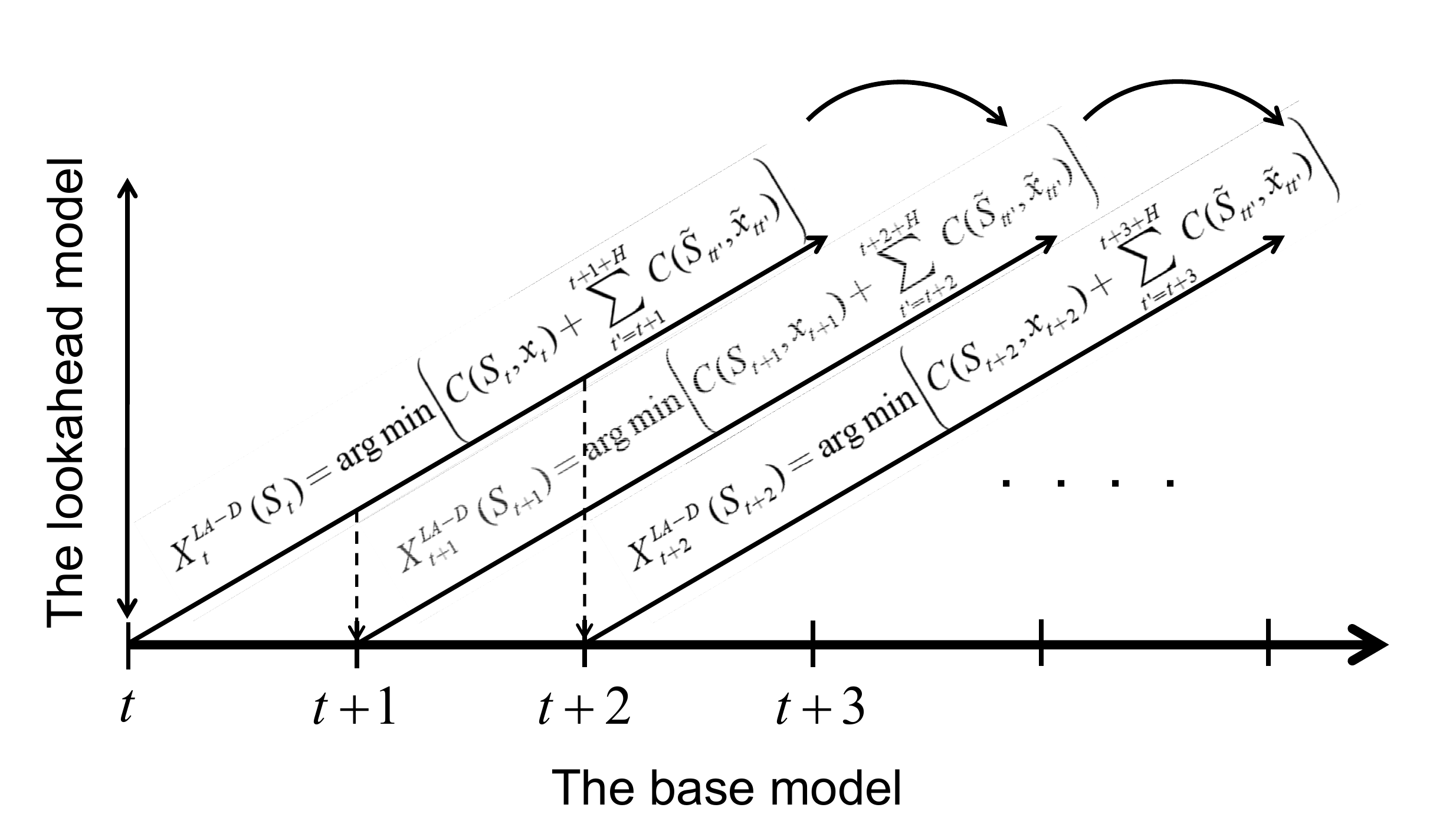}
    \caption{Illustration of simulating a lookahead policy using a deterministic lookahead model.}
    \label{fig:rollingDLA}
\end{center}
\end{figure}


\subsection{Hybrid policies}
A byproduct of identifying the four classes of policies is the ability to create hybrids that combine two or more classes.  Some examples include:
\begin{itemize}
  \item Lookahead policies plus VFAs - We can do an $H$-period lookahead, and then terminate with a value function approximation.  This will likely simplify the task of coming up with a good value function approximation, while also providing better results for the same horizon (allowing us to shorten the horizon for the tree search).
  \item Value function approximations with parameter tuning.  Imagine that we create a VFA-based policy that looks like
      \bn
      X^{VFA}(S_t|\theta) = \argmax_x\left(C(S_t,x) + \sum_{f\in\Fcal} \theta_f \phi_f(S_t,x_t)\right). \label{eq:discussionVFA}
      \en
      Assume we use ADP-based algorithms to determine an estimate of $\theta$.  Now, using this value of $\theta$ as an initial estimate, perform policy search by solving
      \bns
      \max_\theta \sum_{t=0}^T C(S_t,X^{VFA}(S_t|\theta)).
      \ens
      A nice illustration of this strategy is given in \cite{mahe2013}.  It is quite likely that performing this additional tuning (which can be expensive) will further improve the results.  After performing parameter tuning, we can no longer view the linear term as an approximation of the value of being in a state.  After tuning the policy, this is a form of CFA with cost function correction term.
  \item PFA with anything - A policy function approximation is any mapping of state to action without solving an optimization problem.  PFAs are the simplest and easiest to control, but they cannot solve complex problems.  The remaining three classes of policies are all cost-based, which allows them to be used for much more complex problems (including problems where $x_t$ is a high-dimensional vector).  However, cost-based policies are harder to control.

      It is possible to create a hybrid of a PFA and any cost-based policy.  Assume we are using a VFA-based policy $X^{VFA}(S_t|\theta^{VFA})$ (this could also be a direct lookahead or parametric CFA), which we would write as we did in equation \eqref{eq:discussionVFA}, where we let $\theta^{VFA}$ be the coefficients in the value function approximation.  Now assume we are given some parametric function (a PFA) that we represent using $X^{PFA}(S_t|\theta^{PFA})$.  We can write a hybrid policy using parameter vector $\theta = (\theta^{VFA},\theta^{PFA},\theta^{PFA-VFA})$
      \bn
      X^{VFA-PFA}(S_t|\theta) = \argmax_x\left(C(S_t,x) + \sum_{f\in\Fcal} \theta_f  \phi_f(S_t,x_t) + \theta^{PFA-VFA}\|x-X^{PFA}(S_t|\theta^{PFA})\|\right). \nonumber \\ & & \label{eq:discussionVFAPFA}
      \en
      where $\theta^{PFA-VFA}$ handles the scaling between the norm of the difference between $x$ and the decision suggested by $X^{PFA}(S_t|\theta^{PFA})$ and the rest of the cost-based objective function.
\end{itemize}
These hybrid policies help to emphasize the reason why we need to state the objective (as we did in equations \eqref{eq:stateindependentfinalreward} - \eqref{eq:basemodelrisk}) in terms of optimizing over {\it policies}.

\subsection{Remarks}

The academic literature is heavily biased toward the lookahead classes (VFAs and DLAs).  These offer optimal policies, but computable optimal policies are limited to a very small class of problems: the linear control policy for LQR problems in optimal control, and lookup table value functions for problems with small state and action spaces, and a computable one-step transition matrix.

Approximating the value function has extended this framework, but only to a degree.  Approximate/adaptive dynamic programming and $Q$-learning is a powerful tool, but again, algorithms that have been demonstrated empirically to provide near-optimal policies are rare.  Readers have to realize that just because an algorithm enjoys asymptotic optimality (or attractive regret bounds) does not mean that it is producing near-optimal solutions in practice.

It is our belief that the vast majority of real-world sequential decision problems are solved with policies from the policy search class (PFAs and CFAs).  PFAs have received some attention from the research literature (neural networks, linear/affine control policies).  CFAs, on the other hand, are widely used in practice, yet have received minimal attention in the academic literature.

We would argue that PFAs and CFAs should have a place alongside parametric models in machine learning.  A limitation is that they require a human to specify the structure of the parameterization, but this is also a feature: it is possible for domain experts to use their knowledge of a problem to capture structure.  Most important is that PFAs and CFAs tend to be much simpler than policies based on value functions and lookaheads.  But, {\it the price of simplicity is tunable parameters}, and tuning is hard.

\subsection{Stochastic control, reinforcement learning, and the four classes of policies}
The fields of stochastic control and reinforcement learning both trace their origins to a particular model that leads to an optimal policy.  Stochastic control with additive noise (see equation \eqref{eq:stochasticcontroltransition}) produced an optimal policy from the original deterministic model with a quadratic objective function, given by $u_t = K_t x_t$.  Reinforcement learning owes its origins to the field of Markov decision processes, which also produces an optimal policy for discrete problems, where the one-step transition matrix can be computed (see both \eqref{eq:bellman} and \eqref{eq:bellmansteadystate}).

What then happened to both fields is the realization that these optimal policies can only be used in practice for a fairly narrow range of problems.  For this reason, both communities evolved other strategies which can be viewed as being drawn from each of the four classes of policies we have described.
\begin{description}
\item[Optimal control] The following policies can be found in the optimal control literature:
    \begin{description}
     \item[Policy function approximations] These describe a wide range of simple rules used in every day problems.  Some examples are:
        \begin{itemize}
        \item Buy low, sell high - These are simple rules for buying or selling assets.
        \item $(s,S)$ inventory policies - A widely used policy for inventory control is to place an order when the inventory $R_t < s$, in which case we order $S-R_t$.
        \item Linear control laws - Drawing on the optimal policy $u_t = K_t x_t$ for the LQR problems, the controls community branched into general linear ``control laws'' which we might write as
        \bns
        U^\pi(x_t|\theta) = \sum_{f\in\Fcal} \theta_f \phi_f(x_t),
        \ens
        where someone with domain knowledge needs to choose the features $\phi_f(x_t), ~f\in\Fcal$, after which the parameters $\theta$ need to be tuned.
        \end{itemize}
    \item[Value function approximation] The optimal control literature was using neural networks to approximate ``cost-to-go'' functions since the 1970's \citep{We74}. This strategy has been pursued in the controls literature under names including heuristic/neuro/ approximate/adaptive dynamic programming. See \cite{SiBaPo04} for a nice summary of this line of research.
    \item[Direct lookahead] The controls community has referred to policies based on optimizing over a planning horizon as {\it model predictive control}. The most common strategy is to solve deterministic lookahead models.  Most optimal control problems are deterministic, but using a deterministic approximation of the future is typical even when the underlying problem is stochastic (see \cite{Camacho2004} and \cite{rossiter2017} for thorough introductions).
    \item[Parameterized MPC] Some authors in model predictive control have realized that you can obtain better results by introducing parameters as we did in our energy storage problem to handle the uncertainty in forecasts.  This work has been done under the umbrella of ``robust MPC'' (see \cite{Kothare1996} and  \cite{Rakovic2012}).  In our framework, this would be a hybrid direct lookahead-cost function approximation.
    \end{description}
\item[Reinforcement learning] - The following policies are all contained in \cite{Sutton2018}:
    \begin{description}
      \item[Policy function approximation] A popular policy for discrete action spaces is to choose an action based on the Boltzmann distribution given by
          \bns
          p(a|\theta,s) = \frac{e^{\theta \mubar_a}}{1+\sum_{a'\in\Acal_s}e^{\theta \mubar_{a'}}}
          \ens
          where $\mubar_a$ is the current estimate (contained in the state $s$) of the value of action $a$.  The policy is parameterized by $\theta$, which can be optimized using several methods, one of which is known as the ``policy gradient method.''

          In addition to using the policy gradient method on a Boltzmann policy, a number of papers approximate the policy with a neural network.  If $\theta$ is the weights of the neural network, then we have a high-dimensional parameter search problem that we can approach using stochastic gradient algorithms (see, e.g. \cite{spall2003}), although the problem is not easy; simulating policies is noisy, and the problem is not convex.
      \item[Cost function approximation] Upper confidence bounding (for multiarmed bandit problems) is a classic CFA:
          \bns
          X^{CFA}(S^n|\theta) = \argmax_a \left(\mubar^n_a + \theta \sqrt{\frac{\ln n}{N^n_a}}\right)
          \ens
          where $N^n_a$ is the number of times we have tried action $a$ after $n$ iterations.  UCB policies enjoy nice regret bounds \citep{Bubeck2012}, but it is still important to tune $\theta$.
      \item[VFA-based policy] This would be $Q$-learning, where the policy is given by
      \bns
      X^{VFA}(S^n) = \argmax_a \Qbar^n(S^n,a)
      \ens
      \item[Direct lookaheads] Monte Carlo tree search is a classic direct lookahead.  Since MCTS is a stochastic lookahead, it has a policy within the lookahead policy.  This policy looks like
          \bn
          {\tilde X}^\pi_{tt'}(\Stilde_{tt'}|\theta^{UCT}) = \argmax_{\xtilde_{tt'}} \left(\Ctilde(\Stilde_{tt'},\xtilde_{tt'}) + {\tilde V}^x_{tt'}(\Stilde^x_{tt'}) + \theta^{UCT} \sqrt{\frac{\log{N^n_x}}{N^n(\Stilde_{tt'},\xtilde_{tt'})}}\right)
          \en \label{eq:MCTSlookahead}
          Note that this lookahead policy uses both a value function approximation as well as a bonus term from upper confidence bounding.  This logic is known as ``upper confidence bounding on trees,'' abbreviated UCT.  Thus, this is a hybrid policy (CFA with VFA) within a stochastic lookahead.
    \end{description}
\end{description}

So, we see that both the optimal control and reinforcement learning communities are actively using strategies drawn from all four classes of policies.  The same evolution has happened in the simulation-optimization community and the multi-armed bandit community. In the case of multi-armed bandit problems, there are actually distinct communities pursuing the different classes of policies:
\begin{itemize}
\item PFAs - Random sampling of experiments would constitute a PFA (this is a default policy, often used implicitly in statistics).
\item CFAs - Upper confidence bounding is a popular policy for bandit problems in computer science \citep{Bubeck2012}.
\item VFAs - The applied probability community has long used a decomposition technique to produce a series of dynamic programs which can be solved (one ``arm'' at a time) to obtain Gittins indices \citep{gittins2011}.
\item DLAs - These include expected improvement, knowledge gradient and kriging developed in applied math, operations research and geosciences (see the tutorial in \cite{PoFr08} or the book \cite{Powell2012i} for overviews).
\end{itemize}

We now return to our energy storage problem.

\section{Policies for energy storage}
\label{sec:policiesforenergystorage}
We can illustrate all four classes of policies using our energy storage problem.  Note that fully developing any of these classes would require a serious effort, so these are going to be little more than brief illustrations.

\begin{description}
\item[Policy function approximation] - As a simple illustration of a policy function approximation, consider a buy-low, sell-high policy for handling the charging and discharging of a battery connected to the grid.  This policy would be written
    \bn
    X^{PFA}(S_t|\theta) &=& \left\{\begin{tabular}{cc} +1 & $p_t < \theta^{charge}$ \\
                                                        0 & $\theta^{charge} \leq p_t \leq \theta^{discharge}$ \\
                                                        -1 & $p_t > \theta^{discharge}$   \end{tabular}\right. \label{eq:energyPFA}
    \en
    Figure \ref{fig:energyPFA} shows an example of a parameterized PFA for the energy system in figure \ref{fig:energysystem}, where we have highlighted four tunable parameters.  Designing these policies (especially the one in figure \ref{fig:energyPFA}) is an art that requires an understanding of the structure of the problem.  Tuning is an algorithmic exercise.
    \begin{figure}[tb]
    \begin{center}
    \includegraphics[width=4.0in]{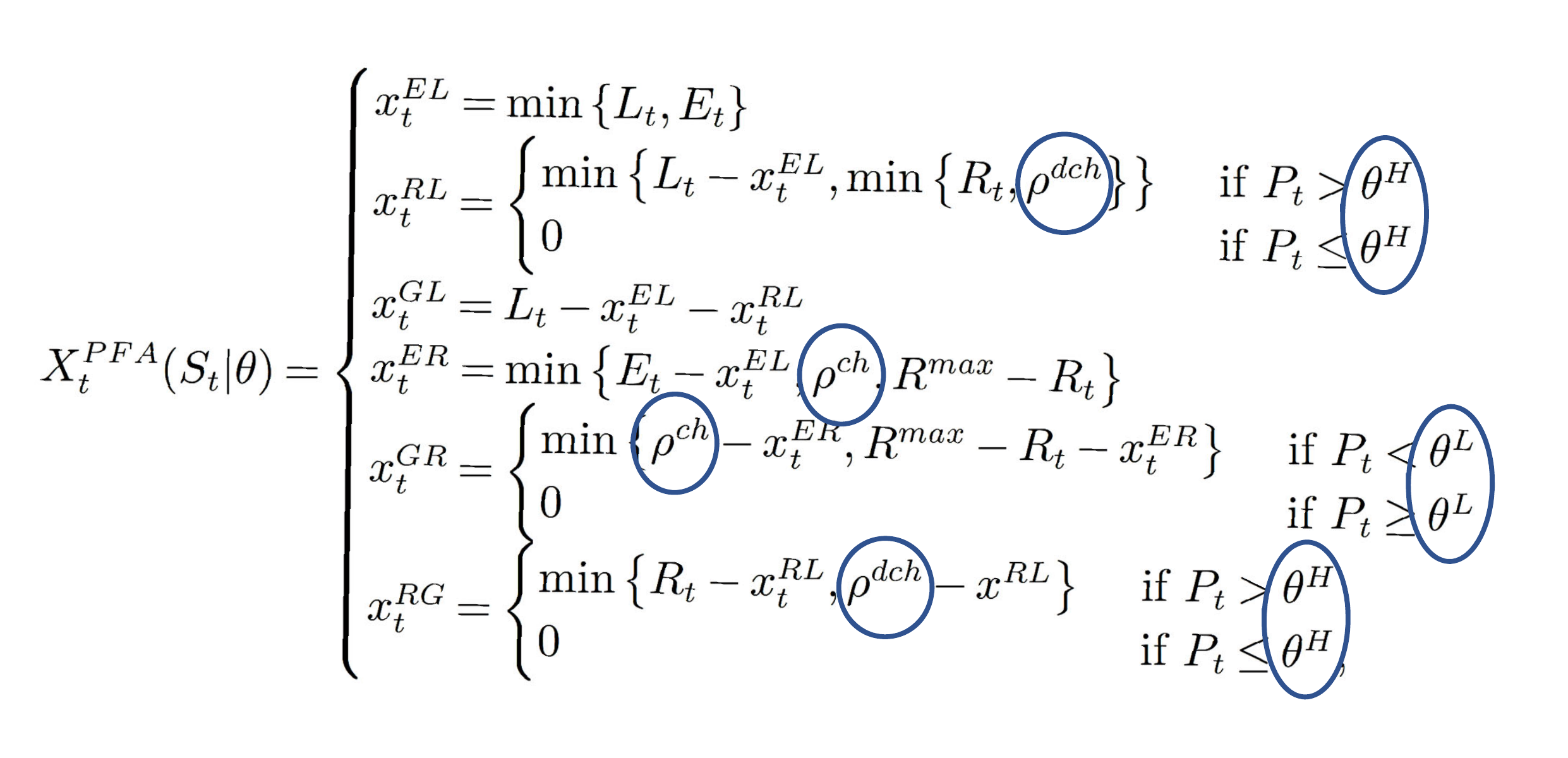}
    \caption{Parameterized policy function approximation for energy system in figure \ref{fig:energysystem}, with tunable parameters highlighted.}
    \label{fig:energyPFA}
    \end{center}
    \end{figure}

    It is important to recognize that we have written our PFAs using parameters $\theta$ that do not depend on time $t$, yet it is clear that in such a time-dependent setting (due to time-of-day patterns and rolling wind forecasts), the policy should be time dependent.  However, tuning a two (or four) dimensional parameter vector $\theta$ is much easier than tuning a time-dependent parameter vector $(\theta_\tau)_{\tau = 1}^{24}$.
\item[Cost function approximation] We are going to present a hybrid cost function approximation with a direct lookahead below.
\item[Value function approximation] We apply the VFA-based policy in equation \eqref{eq:discussionVFA} with the policy
    {\small
    \bns
    X^{VFA}(S_t|\theta) = \argmax_x\left(C(S_t,x) + \big(\theta_1 R_t + \theta_2 R^2_t + \theta_3(x^{EB}_t + x^{ED}_t)^2 + \theta_4(x^{ED}_t + x^{BD}_t + x^{GD}_t)^2\big) \right).
    \ens
    }
    There are a variety of strategies for fitting the coefficients $\theta$ that have been developed under headings of reinforcement learning  \citep{Sutton2018}, approximate dynamic programming (see e.g. \cite{Po11}) and adaptive dynamic programming \citep{SiBaPo04}.  \cite{JiPhPo14} describes an extensive series of tests using value function approximations, where we found that VFA-based policies only worked well when we could exploit structure (such as concavity).
\item[Direct lookahead] For time-varying problems with a rolling forecast, a natural choice is to do a deterministic lookahead.  We do this by setting up a time-staged linear programming model to optimize all decisions over a planning horizon.  This is a deterministic lookahead model, so we let the decisions in the lookahead model created at time $t$ be represented by $\xtilde_{tt'}$, which produces
    \bn
    X^{DLA}(S_t) & = & \argmax_{x_t,(\xtilde_{tt'},t'=t+1, \ldots, t+H)} \left(p_t (x^{GB}_t + x^{GD}_t) + \sum_{t'=t+1}^{t+H} \ptilde_{tt'} (\xtilde^{GB}_{tt'} + \xtilde^{GD}_{tt'}) \right) \label{eq:DLA0}
    \en
    subject to the following constraints. First, for time $t$ we have:
    \bn
    x^{BD}_{t} - x^{GB}_{t} - x^{EB}_{t} & \leq & R_{t}, \label{eq:DLA1} \\
    \Rtilde_{t,t+1}- (x^{GB}_{t} + x^{EB}_{t} - x^{BD}_{t}) &=& R_{t}, \label{eq:DLA2}\\
    x^{ED}_{t} + x^{BD}_{t} + x^{GD}_{t} & = & D_{t},\label{eq:DLA3}\\
    x^{EB}_{t} + x^{ED}_{t}  & \leq & E_{t}, \label{eq:DLA4} \\
    x^{GD}_t, x^{EB}_t, x^{ED}_t, x^{BD}_t & \geq & 0. \label{eq:DLA5}
    \en
    Then, for $t' = t+1, \ldots, t+H$ we have:
    \bn
    \xtilde^{BD}_{tt'} - \xtilde^{GB}_{tt'} - \xtilde^{EB}_{tt'} & \leq & \Rtilde_{tt'}, \label{eq:DLA5} \\
    \Rtilde_{t,t'+1}- (\xtilde^{GB}_{tt'} + \xtilde^{EB}_{tt'} - \xtilde^{BD}_{tt'}) &=& \Rtilde_{tt'}, \label{eq:DLA6}\\
    \xtilde^{ED}_{tt'} + \xtilde^{BD}_{tt'} + \xtilde^{GD}_{tt'} & = & f^D_{tt'},\label{eq:DLA7}\\
    \xtilde^{EB}_{tt'} + \xtilde^{ED}_{tt'}  & \leq & f^E_{tt'}. \label{eq:DLA8}
    \en
\item[Hybrid DLA-CFA] The policy defined by the lookahead model given by equations \eqref{eq:DLA0} - \eqref{eq:DLA8} does not make any provision for handling uncertainty.  The most significant source of uncertainty is the forecast of wind, which is represented deterministically in equation \eqref{eq:DLA8}.  One idea is to parameterize this constraint by replacing it with
    \bn
    \xtilde^{EB}_{tt'} + \xtilde^{ED}_{tt'}  & \leq & \theta_{t'-t} f^E_{tt'}. \label{eq:DLA8a}
    \en
    Now we would write the policy as $X^{DLA-CFA}(S_t|\theta)$ where $\theta = (\theta_\tau)_{\tau=1}^H$ is a set of coefficients for a rolling set of forecasts over a horizon of length $H$.  It is very important to note that $\theta$ is not time-dependent, which means that a policy that needs to behave differently at different times of day becomes a stationary policy, because the forecasts capture all the time dependent information, and the forecasts are captured in the state variable.
\end{description}

The policies in the policy search class, given by $X^{PFA}(S_t|\theta)$ in equation \eqref{eq:energyPFA}, and $X^{DLA-CFA}(S_t|\theta)$ using the parameterized constraint \eqref{eq:DLA8a}, both need to be tuned by solving
\bn
\max_\theta F^\pi(\theta) = \E_{S_0} \E_{W_1, \ldots, W_T|S_0} \sum_{t=0}^T C(S_t,X^\pi(S_t|\theta)), \label{eq:policysearchtheta}
\en
where $S_t$ is governed by the appropriate system model as illustrated in section \ref{sec:energystorage}, and associated information process.

It is unlikely that anyone would test all four classes of policies to see which is best.  A notable exception is \cite{PowellMeisel2016} which showed that any of the four classes of policies (or a hybrid) can work best by carefully choosing the data. It is important to realize that the four classes of policies are meta-classes: simply choosing a class does not mean that your problem is solved.  Each class is actually a path to an entire range of strategies.

\section{Extension to multiagent systems}
\label{sec:multiagent}
We can easily extend our framework to multiagent systems by using the framework to model the environment associated with each agent.  Let $\Qcal$ be the set of agents and let $q\in\Qcal$ represent a specific agent.  There are four types of agents:
\begin{itemize}
  \item The ground truth agent - This agent cannot make any decisions, or perform any learning (that is, anything that implies intelligence).  This is the agent that would know the truth about unknown parameters that we are trying to learn, or which performs the modeling of physical systems that are being observed by other agents.  Controlling agents are, however, able to change the ground truth.
  \item Controlling agents - These are agents that make decisions that act on other agents, or the ground truth agent (acting as the environment).  Controlling agents may communicate information to other controlling and/or learning agents.
  \item Learning agents - These agents do not make any decisions, but can observe and perform learning (about the ground truth and/or other controlling agents), and communicate beliefs to other agents.
  \item Combined controlling/learning agents - These agents perform learning through observations of the ground truth or other agents, as well as making decisions that act on the ground truth or other agents.
\end{itemize}

Now take every variable in our framework and introduce the index $q$.  So, $S_{tq}$ would be the state of the system for agent $q$ at time $t$, which includes:
\bns
R_{tq} &=& \textwrap{The state of resources controlled by agent $q$ at time $t$.}\\
I_{tq} &=& \textwrap{Any other information known to agent $q$ at time $t$.}\\
B_{tq} &=& \textwrap{The beliefs of agent $q$ about anything known to any other agent (and therefore not known to agent $q$).  This covers parameters in the ground truth, anything known by any other agent (for example, the resources that an agent $q'$ might be controlling), and finally, beliefs about how other agents make decisions.}
\ens
Belief states are the richest and most challenging dimension of multiagent systems, especially when there is more than one controlling agent, as would occur in competitive games.

Decisions for agent $q$ are represented by $x_{tq}$.  In addition to decisions that act on the environment, decisions in multiagent systems can include both information collection and communication to other controlling and/or learning agents.  Exogenous information arriving to agent $q$ would be given by $W_{tq}$.  The exogenous information may be observations of a ground truth, or decisions made by other agents.  The transition function gives the equations for updating $S_{tq}$ from decision $x_{tq}$ and exogenous information $W_{t+1,q}$.  The objective function captures the performance metrics for agent $q$.

The design of policies are drawn from the same four classes that we have described above.

One of the controlling agents may play the role of a central agent, but in this framework, a ``central agent'' is simply another agent who makes decisions that are communicated to ``field agents'' who then use these decisions in their planning.

There is a tendency in the literature on multiagent systems to work with a ``system state'' $S_t = (S_{tq})_{q\in\Qcal}$.  We would take the position that this is meaningless, since no agent ever sees all this information.  We would approach the modeling of each agent as its own system, with the understanding that a challenge of any intelligent agent is to develop models that help the agent to forecast the exogenous information process $W_{tq}$.  Of course, this depends on the policy being used by the agent.

A careful treatment of the rich problem of multiagent systems is beyond the scope of this chapter.  However, we feel that the modeling of multiagent systems using this approach, drawing on the four classes of policies, opens up new strategies for the modeling and control of distributed systems.

\section{Observations}
\label{sec:observations}
We are not the first to bridge optimal control with reinforcement learning. \cite{Recht2019} highlights recent successes of reinforcement learning in AlphaGo \citep{Fu2017}, and suggests that these methods should be adapted to control problems.  We would argue that both fields have explored methods that could benefit the other, although we note that the controls community introduced the idea of direct lookaheads (model predictive control) in the 1950's (a form of direct lookahead), affine policies in the 1960's (a form of policy function approximation), and value function approximations in the 1970's.  Both communities have addressed ``model-free'' and ``model-based'' settings, and both have explored methods from all four classes of policies (although neither have investigated parametric CFAs in depth).  We think the biggest difference between optimal control and reinforcement learning is the core motivating applications of each field: optimal control grew originally out of continuous problems of controlling physical devices (aircraft, rockets, robots) while reinforcement learning grew out of problems with discrete action spaces.

We close with the following observations:
\begin{description}
\item[1)] The fields of stochastic control and reinforcement learning address sequential decision problems. We feel that this perspective identifies a range of problems that is much wider than the problems that have been traditionally associated with these communities.
\item[2)] There seems to be considerable confusion about the meaning of ``reinforcement learning.'' In the 1990's and early 2000's, reinforcement learning was a method, called $Q$-learning.  Today, it covers the entire range of methods described in section \ref{sec:policies}.  If we accept that the four classes of policies are universal, then it means that reinforcement learning covers {\it any} policy for a sequential decision problem (the same is true of stochastic control).
\item[3)] Parametric CFAs are often derided by the stochastic optimization community (``heuristics,'' ``deterministic approximations'' are often heard), yet are widely used in practice.  Properly designed parametric policies, however, can be surprisingly effective for two reasons: a) the parameterization can capture domain knowledge that is completely ignored with policies based on lookaheads (VFAs or stochastic DLAs), and b) tuning parameters in a realistic, stochastic base model which avoids the different approximations needed in stochastic lookaheads, can capture complex behaviors that would be overlooked using a simplified stochastic lookahead model.
\item[4)] Stochastic optimal control also addresses sequential decision problems, using a more flexible and scalable modeling framework.  We have argued that the control community is also using instances of all four classes of policies.  So, what is the difference between stochastic optimal control and reinforcement learning?
\item[5)] Our universal framework, which draws heavily on the language used by the stochastic control community, broadens the scope of both of these fields to any sequential decision problem, which we would argue is broader than the problem classes considered by either community.  Further, we have drawn on the four classes of policies identified in \cite{PowellEJOR2019} which encompasses all the strategies already being explored by both communities.  Since our classes are general (they are better described as meta-classes), they help guide the design of new strategies, including hybrids.
%
%
\end{description}


\clearpage
\singlespace
\addcontentsline{toc}{section}{References}
\bibliographystyle{agsm}

\end{document}